\documentclass[10pt,twocolumn,letterpaper]{article}

\usepackage[pagenumbers]{cvpr} %

\usepackage{float}
\usepackage{booktabs}
\usepackage{adjustbox}
\usepackage{makecell}
\usepackage{graphicx}
\usepackage{multirow}
\usepackage{comment}
\usepackage{pifont}
\usepackage{tabularx}
\usepackage[ruled,linesnumbered]{algorithm2e}

\usepackage{tcolorbox}
\definecolor{lightroyalblue}{HTML}{F6F8FD} %
\definecolor{royalblue}{HTML}{4169E1}
\definecolor{lighterblue}{HTML}{f2fafd}  %
\newtcolorbox{abox}{colback=lightroyalblue,colframe=black}

\definecolor{myred}{HTML}{F46F68} %
\newcommand{\myred}[1]{\textcolor{myred}{#1}}

\definecolor{cvprblue}{rgb}{0.21,0.49,0.74}
\usepackage[pagebackref,breaklinks,colorlinks,allcolors=cvprblue]{hyperref}

\title{HMGIE: Hierarchical and Multi-Grained Inconsistency Evaluation for Vision-Language Data Cleansing}

\author{
Zihao Zhu\textsuperscript{1}\ \ \ \ 
Hongbao Zhang\textsuperscript{1} \ \ \ \ 
Guanzong Wu\textsuperscript{1} \ \ \ \ 
Siwei Lyu\textsuperscript{2} \ \ \ \ 
Baoyuan Wu\textsuperscript{1}\thanks{Corresponds to Baoyuan Wu (\href{wubaoyuan@cuhk.edu.cn}{wubaoyuan@cuhk.edu.cn}).} \\
\textsuperscript{1}School of Data Science, The Chinese University of Hong Kong, Shenzhen\\
\textsuperscript{2}University at Buffalo, State University of New York\\
}

\begin{document}
\maketitle

\begin{abstract}
Visual-textual inconsistency (VTI) evaluation plays a crucial role in cleansing vision-language data. 
Its main challenges stem from the high variety of image captioning datasets, where differences in content can create a range of inconsistencies (\eg, inconsistencies in scene, entities, entity attributes, entity numbers, entity interactions). Moreover, variations in caption length can introduce inconsistencies at different levels of granularity as well. 
To tackle these challenges, we design an adaptive evaluation framework, called Hierarchical and Multi-Grained Inconsistency Evaluation (HMGIE), which can provide multi-grained evaluations covering both accuracy and completeness for various image-caption pairs. 
Specifically, the HMGIE framework is implemented by three consecutive modules. 
Firstly, the semantic graph generation module converts the image caption to a semantic graph for building a structural representation of all involved semantic items. 
Then, the hierarchical inconsistency evaluation module provides a progressive evaluation procedure with a dynamic question-answer generation and evaluation strategy guided by the semantic graph, producing a hierarchical inconsistency evaluation graph (HIEG).
Finally, the quantitative evaluation module calculates the accuracy and completeness scores based on the HIEG, followed by a natural language explanation about the detection results.
Moreover, to verify the efficacy and flexibility of the proposed framework on handling different image captioning datasets, we construct MVTID, an image-caption dataset with diverse types and granularities of inconsistencies. 
Extensive experiments on MVTID and other benchmark datasets demonstrate the superior performance of the proposed HMGIE to current state-of-the-art methods.
\end{abstract}

\section{Introduction}
\label{sec:intro}

\begin{figure}
    \centering
    \includegraphics[width=\columnwidth]{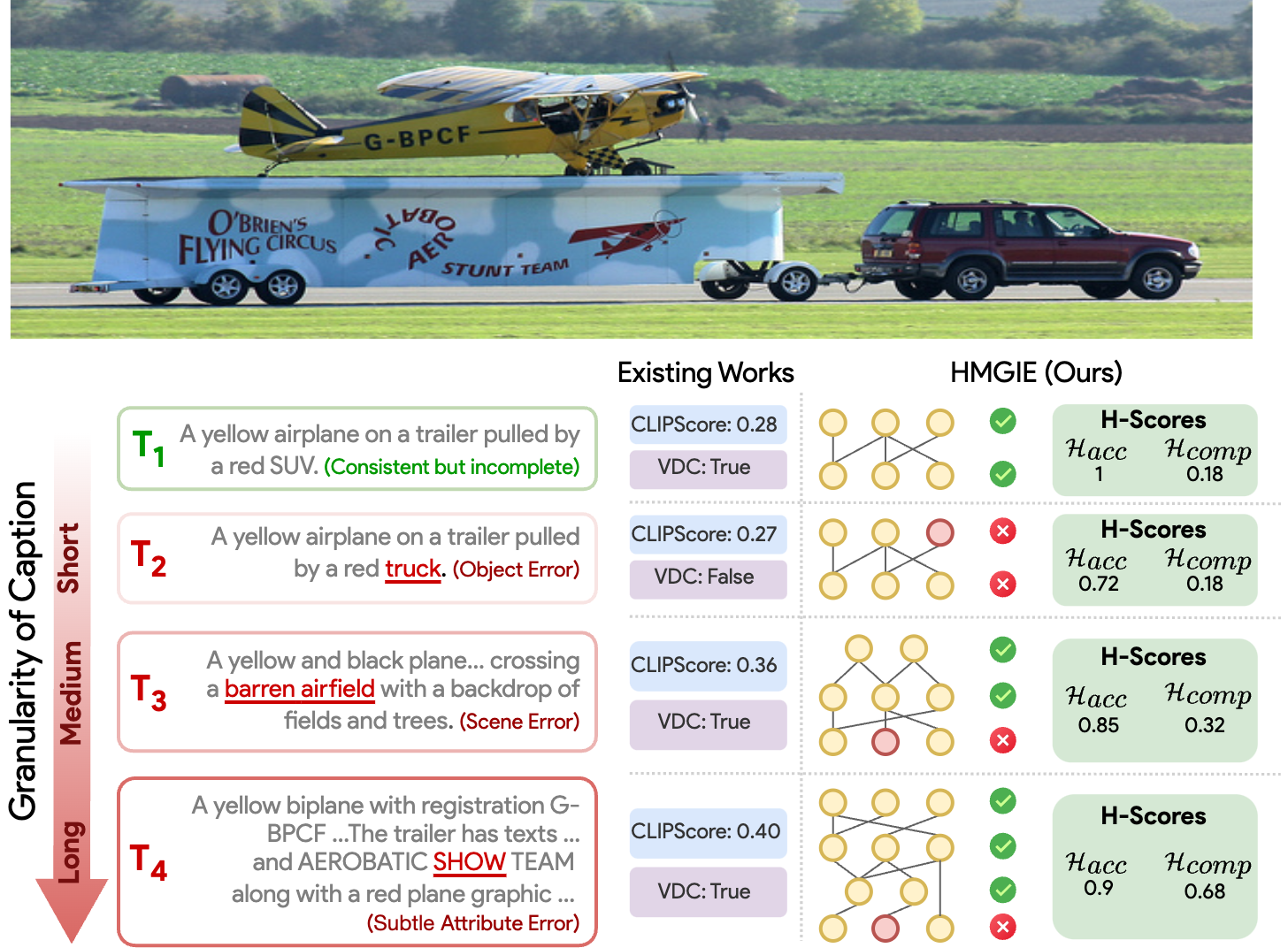}
    \caption{Comparison of inconsistency evaluation methods on captions with various granularities. Captions range from short to detailed, containing different types of inconsistencies (object error in $T_2$, scene error in $T_3$, and attribute error in $T_4$). Existing methods show limitations: CLIPScore fails to differentiate between consistent and inconsistent short captions ($T_1$ \vs $T_2$), while VDC misses inconsistencies in longer captions ($T_3$, $T_4$). In contrast, HMGIE provides more reasonable h-scores: $\mathcal{H}_{acc}$ for semantic accuracy and $\mathcal{H}_{comp}$ for semantic completeness.}
    \vspace{-10pt}
    \label{fig:intro}
\end{figure}

The quality of vision-language data plays a critical role in training vision-language models (VLMs), such as multimodal large language models (MLLMs)~\cite{alayrac2022flamingo,li2023blip,zhuminigpt,ye2024mplug} and text-to-image diffusion models~\cite{epstein2023diffusion,croitoru2023diffusion,zhang2023adding,ruiz2023dreambooth}. However, large-scale vision-language datasets collected from the internet or through human annotation, typically in the form of image-caption pairs, often suffer from \textbf{visual-textual inconsistency} issues, where captions contain semantic discrepancies with their corresponding images~\cite{bai2024survey,qin2024synergy,gadre2024datacomp}. These pervasive data quality issues significantly impact downstream task performance, highlighting the crucial need for automatic visual-textual inconsistency evaluation to facilitate effective vision-language data cleansing. 

In real-world scenarios, as illustrated in Figure~\ref{fig:intro}, image-caption data exhibits significant diversity, presenting two main challenges for inconsistency evaluation. First, captions vary substantially in length and granularity, ranging from concise captions that capture basic scene elements (\eg, $T_1$, $T_2$) to comprehensive narratives that include intricate details (\eg, $T_4$ describing the aircraft's registration number and trailer's text content). Second, inconsistencies show diverse types, including object misidentification (\eg, \textit{``truck"} instead of \textit{``SUV"} in $T_2$), scene mischaracterization (\eg, \textit{``barren airfield"} in $T_3$ contradicting the \textit{grassy field}), and subtle attribute errors (\eg, misrepresenting \textit{``CIRCUS"} as \textit{``SHOW"} in $T_4$). However, existing methods struggle to address these challenges effectively. Embedding-based approaches like CLIPScore~\cite{hessel2021clipscore}, which rely on global image-text similarity, fail to capture local semantic inconsistencies --- assigning nearly identical scores to both consistent $T_1$  and inconsistent $T_2$. Meanwhile, QA-based methods~\cite{vqascore,hu2023tifa,yarom2024you}  show limitations in detecting fine-grained inconsistencies, particularly in detailed captions. For example, VDC~\cite{zhu2023vdc} has been shown to incorrectly validate inconsistent cases in $T_3$ and $T_4$. Therefore, developing an evaluation framework that can effectively handle varying granularities and diverse types of inconsistencies remains an urgent challenge in vision-language data cleansing.

\begin{figure*}[!ht]
    \centering
    \includegraphics[width=\linewidth]{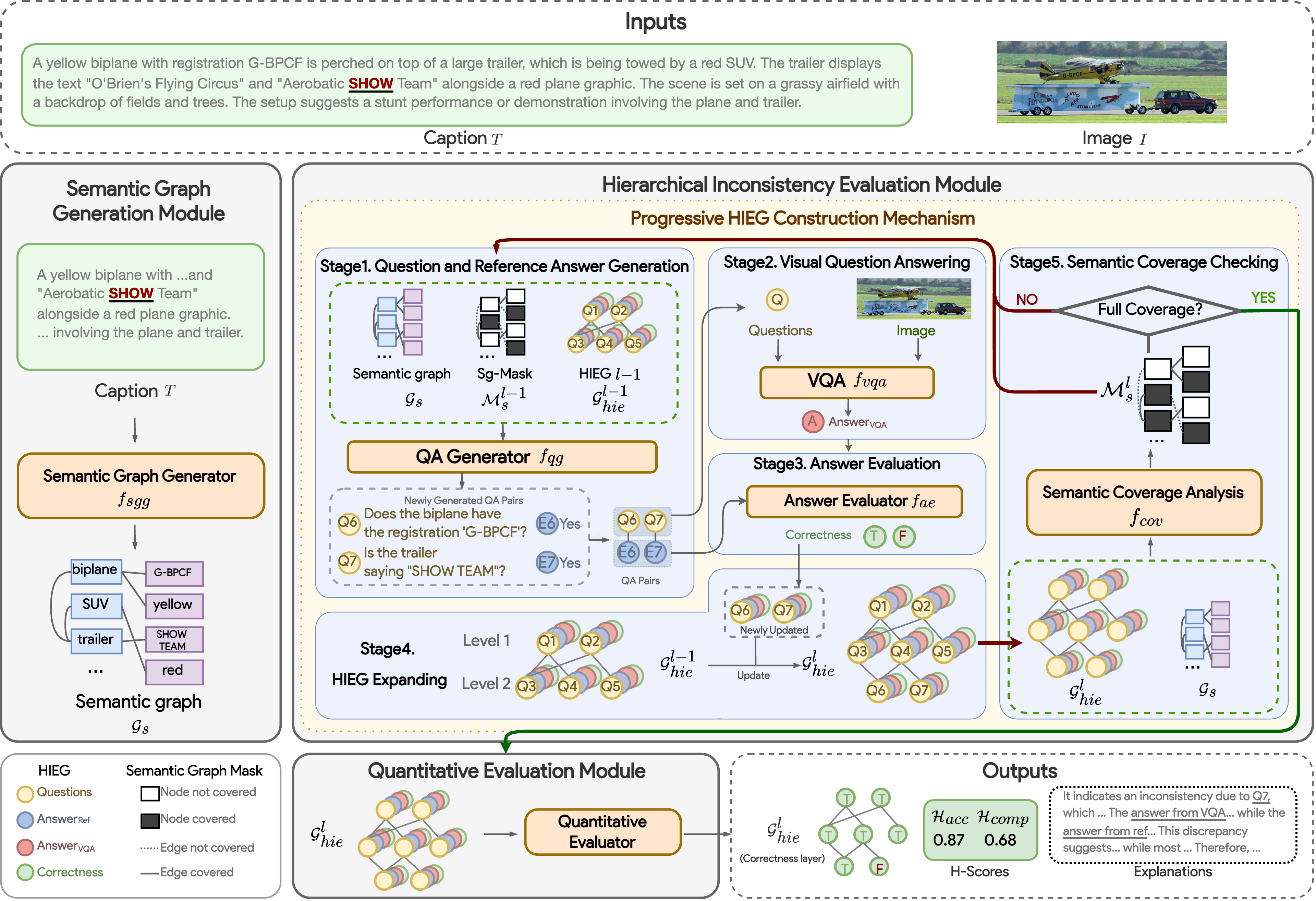}
    \caption{The overall illustration of our proposed hierarchical and multi-grained inconsistency evaluation (HMGIE) framework.}
    \vspace{-6pt}
    \label{fig:framework}
\end{figure*}

To address the challenges, we propose HMGIE, a hierarchical and multi-grained inconsistency evaluation framework that provides adaptive evaluations for multi-grained image-caption data. As shown in Figure \ref{fig:framework}, HMGIE consists of three core modules: (1) a semantic graph generation module that converts captions into semantic graphs, representing structured relationships among semantic elements; (2) a hierarchical inconsistency evaluation module that progressively constructs a hierarchical inconsistency evaluation graph (HIEG) through dynamic question-answer generation and evaluation guided by the semantic graph, followed by semantic coverage checking that tracks examined elements and guides subsequent question generation; and (3) a quantitative evaluation module that calculates complementary H-Scores measuring both semantic accuracies ($\mathcal{H}_{acc}$) and semantic completeness ($\mathcal{H}_{comp}$), and offers overall consistency decisions and convincing natural language explanations. This hierarchical architecture enables HMGIE to achieve practical VTI evaluation across captions of various types and granularity. To facilitate a thorough evaluation of our framework, we construct MVTID, the first multi-grained visual-textual inconsistency dataset that includes four distinct granularities. Extensive experiments on the MVTID dataset and other benchmarks demonstrate that HMGIE significantly outperforms current state-of-the-art VTI evaluation methods.

Our main contributions are \textbf{four-fold}: (1) We propose HMGIE, a novel VTI evaluation framework that can effectively handle multiple granularities and diverse types of inconsistencies in vision-language data.
(2) We introduce complementary H-Scores that provide a comprehensive evaluation by both measuring semantic accuracy and completeness.
(3) We construct MVTID, a challenging multi-granularity dataset with diverse types of inconsistencies across different granularity levels.
(4) Extensive experiments on several benchmarks and human evaluation demonstrate HMGIE's superior performance in inconsistency evaluation.

\section{Related Work}
\label{sec:related}

\subsection{Embedding-based Detection}
Early efforts~\cite{hessel2021clipscore, negclip, ghosh2024geneval, fan2024exploring, hongcas} to detect visual-textual inconsistency focused on learning and comparing cross-modal representations within shared embedding spaces. For example, CLIPScore~\cite{hessel2021clipscore} pioneers this approach, utilizing CLIP's~\cite{clip} pre-trained image and text encoders to calculate cosine similarity between cross-modal embeddings. Building on this foundation, NegCLIP~\cite{negclip} recognizes that vision-language models often act as bags of words and introduces composition-aware contrastive learning to better capture relationships and attributes. However, these embedding-based methods encounter inherent limitations when addressing fine-grained semantic inconsistencies, especially in longer, more complex captions, as global embedding similarity fails to capture the nuanced semantic details necessary for comprehensive consistency evaluation.

\subsection{Prompt-based Detection with MLLMs}
The emergence of multimodal large language models (MLLMs)~\cite{dai2023instructblip, gao2023llama,ye2023mplug,lu2024deepseek} has enabled more sophisticated methods to detect inconsistencies using carefully designed prompts, grouping methods into two main categories:

\noindent\textbf{Direct prompting methods} leverage MLLMs' superior cross-modal understanding capabilities through single-turn prompts to query MLLMs about image-text similarity~\cite{vqascore, zhang2024data,bai2024survey}.
For example, VQAScore~\cite{vqascore} reformulates captions into yes/no questions (\eg, ``Does this figure show \{caption\}?") and uses the model's confidence in responding ``yes" as a consistency score. Even with chain-of-thought~\cite{wei2022chain} reasoning to decompose evaluation steps, these methods still struggle to capture subtle semantic misalignments like incorrect attributes or spatial relationships.

\noindent\textbf{QA-based methods} adopt question-answering frameworks for consistency evaluation. TIFA~\cite{hu2023tifa} and VDC~\cite{zhu2023vdc} decompose captions into multiple targeted questions, assessing consistency by comparing text-based answers against visual question answering results from MLLMs. VQ2~\cite{yarom2024you} builds on this approach by using MLLMs to directly verify the quality of generated QA pairs through visual evaluation. DSG~\cite{JaeminCho2024} introduces question dependencies to enable systematic semantic analysis. While these QA-based methods have proven effective, they generate all questions simultaneously, overlooking the hierarchical nature of semantic evaluation. This design limits their ability to handle captions of multiple granularities.

\section{Methodology}
\label{sec:method}

\subsection{Task Definition and Overview}
\noindent\textbf{Problem Formulation.}
Given an image $I$ and its corresponding caption $T$, we aim to evaluate visual-textual inconsistency (VTI) between them. In real-world applications, image captions exhibit high variety in both content and length, leading to diverse types of inconsistencies (\eg, inconsistencies in scene, entities, attributes, interactions) and varying granularity of inconsistencies. Therefore, comprehensive VTI evaluation requires analyzing both accuracy and completeness across different granularity levels.

\noindent\textbf{Framework Overview.}
As illustrated in Figure~\ref{fig:framework}, we propose a \textbf{H}ierarchical and \textbf{M}ulti-\textbf{G}rained \textbf{I}nconsistency \textbf{E}valuation (HMGIE) framework, which provides adaptive evaluations covering both semantic accuracy and completeness for multi-grained image-caption data. 
The framework consists of three consecutive modules: (1)semantic graph generation module that converts image captions into a semantic graph $\mathcal{G}_{sg}$, representing structured relationships among semantic elements; (2)hierarchical  inconsistency evaluation module that adopts a hierarchical inconsistency evaluation graph (HIEG) $\mathcal{G}_{hie}$ construction mechanism; and (3)quantitative evaluation module that calculates corresponding H-Scores, overall consistency decision, along with natural language explanations. This hierarchical architecture enables HMGIE to achieve effective VTI evaluation across captions of various types and granularity.

\subsection{Semantic Graph Generation Module}
 Unlike existing QA-based methods that generate questions directly from raw text, HMGIE first converts the caption $T$ into a semantic graph $\mathcal{G}_{s} = (\mathcal{V}, \mathcal{E})$, where $\mathcal{V}$ represents the set of nodes and $\mathcal{E}$ denotes the set of edges.
 
\noindent\textbf{Graph Structure Design}.
Each node $v_i \in \mathcal{V}$ represents different semantic elements in the text description, including entity, location, concept, event, attribute, and other auxiliary components. The edges $e_{i,j} \in \mathcal{E}$ connecting nodes $v_i$ and $v_j$ capture various relationships between nodes, including action, spatial relations, attribute connections, part-whole relationships, quantitative associations, and other semantic relationships. This structured representation enables precise semantic understanding and systematic evaluation planning. Detailed examples of semantic graphs can be found in Appendix~\ref{app:experiment}.

\noindent\textbf{Semantic Parsing Process.}
The semantic parsing process $f_{sgg}: T \rightarrow \mathcal{G}_{s}$ is implemented by an LLM with carefully designed prompts that ensure consistent and comprehensive graph construction, following recent advances in LLM-based semantic graph generation~\cite{chen2023gpt4sgg,kim2024llm4sgg}. Given a caption $T$, the model first identifies and assigns all semantic elements to appropriate node types. Then, it analyzes the relationships between these elements to establish edges, considering both explicit connections stated in the text and implicit relationships that can be inferred. The resulting semantic graph provides several key advantages: (1) it captures complex semantic relationships more effectively than linear text, (2) it facilitates targeted question generation by identifying key semantic elements, and (3) it enables semantic coverage tracking through explicit node and edge representations.

\begin{table*}[!t]
\centering
\caption{Specifications and examples of different granularity levels in MVTID. Texts in \textcolor{red}{red} indicate the inaccurate information.}
\label{tab:granularity}
\renewcommand{\arraystretch}{0.5}  %
\resizebox{0.9\textwidth}{!}{
\begin{tabular}{>{\centering\arraybackslash}p{1.5cm} >{\centering\arraybackslash}p{5cm} >{\centering\arraybackslash}p{2cm} >{\centering\arraybackslash}c >{\centering\arraybackslash}p{10cm}}
\toprule
Granularity & \makecell{Content Requirements} & \makecell[c]{Avg Words \\ ($\pm$ std)} & \makecell[c]{Image} & \makecell[c]{Noisy Caption Example} \\
\midrule
\makecell[l]{G1:\\ Basic} & \makecell[l]{ • Main objects identification\\ • Simple sentence structure\\ • Basic scene type} & \makecell[c]{11.09 \\ $\pm$ 2.02} & $\raisebox{-.5\height}{\includegraphics[width=0.1\linewidth]{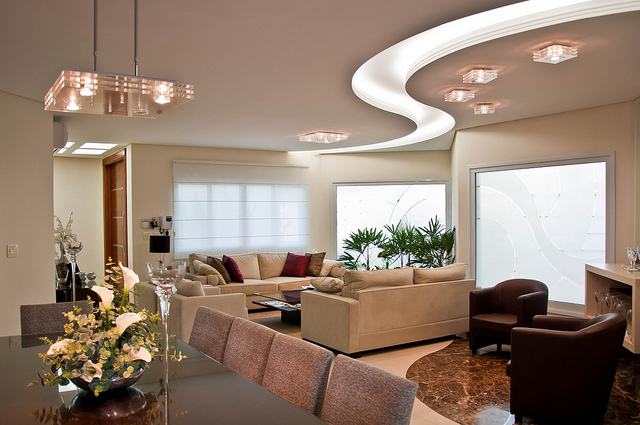}}$ & \makecell[l]{\small{A modern living room with sofas, a dining table \textcolor{red}{and pink flowers}.}} \\
\midrule

\makecell[l]{G2:\\ Extended} & \makecell[l]{  • Basic visual attributes\\ • Object locations\\ • Primary spatial relationships} & \makecell[c]{24.41 \\ $\pm$ 4.30} & $\raisebox{-.5\height}{\includegraphics[width=0.1\linewidth]{fig/example_img.jpg}}$ & 
\makecell[l]{
\small{The image shows a living room with beige sofas near \textcolor{red}{a fireplace}. There are }\\
\small{brown armchairs and a dining table with chairs. The room features a distinctive }\\
\small{curved ceiling.}
}
\\
\midrule
\makecell[l]{G3:\\ Detailed}  & \makecell[l]{ • Secondary objects\\ • Specific visual features\\ • Complex relationships} & \makecell[c]{40.93 \\ $\pm$ 6.63} & $\raisebox{-.5\height}{\includegraphics[width=0.1\linewidth]{fig/example_img.jpg}}$ & 
\makecell[l]{\small{The living room features a beige L-shaped sofa set with cushions and \textcolor{red}{light} brown}\\
\small{\textcolor{red}{wooden chairs}. Frosted windows allow light into the space, which is illuminated by a}\\
\small{combination of recessed and floor lamps. Marble flooring adds a touch of elegance,}\\
\small{and floral arrangements enhance the decor. The ceiling has a modern, curved design.}
}
\\
\midrule
\makecell[l]{G4:\\ Complete} & \makecell[l]{• All visual elements\\ • Full attribute descriptions\\ • Comprehensive scene details} & \makecell[c]{63.88 \\ $\pm$ 9.05} & $\raisebox{-.5\height}{\includegraphics[width=0.1\linewidth]{fig/example_img.jpg}}$ & 
\makecell[l]{\footnotesize{The image depicts a modern living room featuring a large beige sofa with \textcolor{red}{blue pillows},}  \\ 
\footnotesize{accompanied by dark brown armchairs. The room is illuminated by recessed square lighting}\\ 
\footnotesize{fixtures in a curved ceiling design. Frosted glass windows allow natural light, while marble}\\ 
\footnotesize{flooring adds elegance. Decorative floral arrangements enhance the contemporary aesthetic.}\\
\footnotesize{The neutral color palette of beige and brown is consistent throughout the space.}
}\\
\bottomrule
\end{tabular}%
}
\vspace{-5pt}
\end{table*}

\subsection{Hierarchical Inconsistency Evaluation Module}
Given the semantic graph $\mathcal{G}_{s}$, this module progressively constructs a hierarchical inconsistency evaluation graph (HIEG) through dynamic question-answer generation and evaluation guided by the semantic graph, followed by semantic coverage checking.

\noindent\textbf{HIEG Definition.}
Let $\mathcal{G}_{hie} = (\mathcal{N}, \mathcal{R})$ denote our defined {\it Hierarchical Inconsistency Evaluation Graph}, where $\mathcal{N}$ represents the set of all evaluation nodes across different levels, and $\mathcal{R}$ captures their hierarchical dependencies. Each node $n_i^l \in \mathcal{N}$ at level $l$ represents an evaluation unit consisting of a quartet $(q_i^l, a_{ref, i}^l, a_{vqa, i}^l, y_i^l)$, where $q_i^l$ is the generated question, $a_{ref,i}^l$ denotes the reference answer derived from semantic graph, $a_{vqa,i}^l$ represents the answer generated by visual question answering, and
$y_i^l \in \{0,1\}$ indicates whether $a_{vqa,i}^l$ correctly aligns with $a_{ref,i}^l$.
The node set $\mathcal{N}$ is organized hierarchically across $L$ evaluation levels: $\mathcal{N} = \cup_{l=1}^L \mathcal{N}^l$, where $\mathcal{N}^l$ denotes the set of nodes at level $l$. The dependency relationship set $\mathcal{R}$ is defined as:
\begin{equation}
\mathcal{R} = \{(n_i^k, n_j^l) | n_i^k \in \mathcal{N}_k, n_j^l \in \mathcal{N}_l, k < l \leq L\},
\end{equation}
where $(n_i^k, n_j^l)$ indicates that the evaluation node $n_j^l$ at level $l$ depends on node $n_i^k$ from any previous level $k$, which enables the generation of more specific and targeted questions by building upon previously asked questions, allowing for progressive refinement of the evaluation process.

\noindent\textbf{Progressive HIEG Construction Mechanism.}
The HIEG is constructed progressively through iterative expansion at each level. At level $l$, the construction process involves:

\ding{169} \textbf{Stage 1: Question and Reference Answer Generation.} 
Given the semantic graph $\mathcal{G}_{s}$, a binary mask $\mathcal{M}^{l-1}$ of the $\mathcal{G}_{s}$ indicating unverified elements from previous level, and the existing HIEG $\mathcal{G}_{hie}^{l-1}$ up to level $l-1$, the question generation module produces a set of insightful questions $\mathcal{Q}^l = \{q_1^l, ..., q_{n_l}^l\}$ for level $l$, their corresponding reference answers $\mathcal{A}_{ref}^l = \{a_{ref,1}^l, ..., a_{ref,n_l}^l\}$, and dependencies on existing questions $\mathcal{R}^l$:
\begin{equation}
(\mathcal{Q}^l, \mathcal{A}_{ref}^l, \mathcal{R}^l) = f_{qg}(\mathcal{G}_{s}, \mathcal{M}^{l-1}, \mathcal{G}_{hie}^{l-1}, \pi_l),
\end{equation}
where $f_{qg}$ leverages LLM's capabilities with carefully designed prompts that incorporate predefined level-specific question generation guidelines $\pi_l$. These guidelines direct model to generate questions with increasing complexity: from basic object identification at early levels to subtle semantic nuances at deeper levels. This progressive refinement ensures systematic coverage from coarse-grained to fine-grained semantic elements.

\ding{169} \textbf{Stage-2: Visual Question Answering.} 
For each generated question $q_i^l$, the actual answer $a_{vqa,i}^l$ is generated by reasoning on the input visual content:
$
(a_{vqa,i}^l, c_i^l) = f_{vqa}(I, q_i^l),
$
where $f_{vqa}$ represents the visual question answering function that can be implemented by any VQA model or MLLM, and $c_i^l \in [0,1]$ indicates the model's confidence in its answer. This step extracts semantic information implied in the image through visual reasoning, producing answers that reflect the actual visual content.

\ding{169} \textbf{Stage-3: Answer Evaluation.} Then the correctness of each vqa-generated answer is then evaluated against the corresponding reference answer:
$
y_i^l = f_{eval}(a_{vqa,i}^l, a_{ref,i}^l),
$
where $f_{eval}$ is implemented by determining semantic equivalence rather than exact matching, and $y_i^l \in \{0,1\}$ indicates whether the question is correctly answered.

\ding{169} \textbf{Stage-4: HIEG Expanding.} Based on the generated questions, reference and actual answers, and evaluated results, new nodes are added to the HIEG and connected according to their dependencies:
\begin{equation}
\mathcal{G}_{hie}^l = \mathcal{G}_{hie}^{l-1} \cup \{(q_i^l, a_{ref,i}^l, a_{vqa,i}^l, y_i^l)\}_{i=1}^{n_l} \cup \mathcal{R}^l,
\end{equation}
This expansion process maintains the hierarchical structure of the evaluation process while capturing the semantic relationships among questions at different levels.

\ding{169} \textbf{Stage-5: Semantic Coverage Checking.}
After expanding the HIEG at level $l$, HMGIE analyzes which semantic elements have been examined through tracking the coverage of $\mathcal{G}_{s}$ within the $\mathcal{G}_{hie}^l$:
\begin{equation}
\mathcal{M}^l = f_{cov}(\mathcal{G}_{s}, \mathcal{G}_{hie}^l),
\end{equation}
where $f_{cov}$ tracks the status of each semantic element by analyzing which nodes and edges in $\mathcal{G}_{s}$ have been examined by questions in $\mathcal{G}_{hie}^l$. The output $\mathcal{M}^l$ is a binary mask with the same structure as $\mathcal{G}_{s}$ where 1 indicates unexamined elements, which guide question generation in subsequent levels. 
If $\mathcal{M}^l$ contains any unexamined elements and the maximum level $K$ has not been reached, the framework iterates back to stage 1  to generate questions for the next level $l+1$.

\subsection{Quantitative Evaluation Module}
Based on the fully constructed HIEG with $L$ levels $\mathcal{G}^L_{hie}$, we perform a comprehensive quantitative evaluation that consists of three complementary components: H-Scores computation, overall consistency assessment, and natural language explanation generation.

\noindent\textbf{H-Scores Computation.}
To provide a more nuanced evaluation of both semantic consistency and completeness, we introduce two complementary metrics derived from HIEG, collectively referred as H-Scores, including the accuracy score and the completeness score:
The accuracy score $\mathcal{H}_{acc}$ quantifies the degree of semantic consistency across all evaluation levels:
\begin{equation}
\mathcal{H}_{acc} = \sum_{j=1}^{L} \omega_j \left(\frac{1}{n_j} \sum_{i=1}^{n_j} c_i^j y_i^j\right), \text{ s.t. } \sum_{j=1}^{L}\omega_j=1,
\end{equation}
where $\omega_j$ represents level-specific weights that sum to 1, $c_i^j$ is the confidence score of the answer, and $y_i^j$ is the correctness indicator for each question-answer pair. 
The completeness score $\mathcal{H}_{comp}$ measures the semantic richness of the caption by measuring the amount of evaluation content:
\begin{equation}
\mathcal{H}_{comp} = \sum_{j=1}^{L} \alpha_j \frac{n_j}{N_j}, \text{ s.t. } \sum_{j=1}^{K}\alpha_j=1,
\end{equation}
where $n_l$ denotes the number of evaluation nodes at level $l$, $N_l$ represents the maximum allowable nodes at that level, $K$ represents the maximum allowed evaluation depth, and $\alpha_l$ are level-specific weights. A higher completeness score indicates more comprehensive semantic content requiring extensive evaluation.

\noindent\textbf{Overall Consistency Assessment.}
The overall consistency of $\mathcal{G}^L_{hie}$ is calculated by evaluating the correctness of all evaluation nodes across all levels:
$
d^L = \prod_{l=1}^L \prod_{i=1}^{n_l} y_i^l \in \{0,1\},
$
where $n_l$ denotes the number of nodes at level $l$, and $y_i^l$ represents the correctness indicator for the $i$-th node at level $l$. It means that the image-caption pair is considered consistent only if all questions are answered correctly across all levels of HIEG. This binary output will be used to calculate the conventional metrics (\eg, TPR, FPR, specified later), to ensure the comparison with other inconsistency evaluation methods which can only give binary outputs.

\noindent\textbf{Explanation Generation.}
To enhance interpretability, HMGIE generates natural language explanations by analyzing the evaluation path through the HIEG:
$\xi_l = f_{exp}(\mathcal{G}_{hie}^l, d)$,
where $f_{exp}$ leverages a LLM to synthesize evaluation results into coherent explanations. The explanation highlights critical inconsistent elements, offering users clear insights into the evaluation process and results.

\section{MVTID Dataset}
\label{sec:dataset}

Existing VTI datasets~\cite{levinboim2021quality,zhu2023vdc,yu2023delving} primarily focus on short-length captions or single-label, limiting comparison evaluation where captions vary significantly in granularity. To address this limitation, we introduce  the first \textit{Multi-granularity Visual-Textual Inconsistency Dataset (MVTID)}.

\subsection{Adversarial Dataset Construction}
We employ an adversarial framework that leverages LLMs to generate challenging inconsistent samples at multiple granularities, which consists of following two stages:

\noindent\textbf{Ground-truth Caption Generation.} 
For each image $I_i$ randomly selected from the COCO dataset~\cite{chen2015microsoft}, we generate ground-truth captions $\mathcal{T}_i = \{T_{i,1}, ..., T_{i,4}\}$ at four distinct granularities using an ensemble approach. Firstly, three different MLLMs generate initial captions:
$
\hat{T}_{i,j}^k = f_k(I_i, g_j),
$
where $f_k$ represents the $k$-th MLLM and $g_j$ denotes the granularity-specific generation prompt. These captions are then fused using an LLM to create the final ground-truth caption $T_{i,j}$:
$
T_{i,j} = f_s(\{\hat{T}_{i,j}^k\}), 
$
where $f_s$ identifies and preserves consistent semantic elements across the three generated captions while ensuring coherence and naturalness.

\noindent\textbf{Adversarial Perturbation.}
For each ground-truth caption $T_{i,j}$, we generate its adversarial counterpart through an iterative perturbation process. At iteration $t$, we generate a perturbed caption:
\begin{equation}
\tilde{T}_{i,j}^t = h(T_{i,j}, \{\tilde{T}_{i,j}^k\}_{k=0}^{t-1}),
\end{equation}
where $h$ represents the perturbation function implemented by an LLM that considers all previous perturbations. Then, another MLLM evaluates the inconsistency of the perturbed caption. This process continues iteratively until either the inconsistency becomes undetectable or the maximum iteration count is reached.

As shown in Table~\ref{tab:granularity}, MVTID encompasses four distinct granularities, each providing progressively richer semantic information. Beginning with basic descriptions (G1) that capture primary objects and scene categories and advancing to complete narratives (G4) that comprehensively detail all visual elements and their interrelationships, the average caption length increases from 11.09 to 63.88 words. This structured progression in granularity enables a thorough evaluation of inconsistency detection methods across various levels of semantic complexity. More examples are presented in Appendix~\ref{app:dataset}.

\begin{table*}[htbp]
\centering
\caption{Comparison of visual-textual inconsistency detection results (TPR, FPR, and F1 in \%) across different granularity on the MVTID.}
\label{tab:res_main}
\renewcommand\arraystretch{1}
\resizebox{0.95\textwidth}{!}{%
\begin{tabular}{@{}lccccccccccccccc@{}}
\toprule
Granularity$\rightarrow$ & \multicolumn{3}{c}{G1: Basic} & \multicolumn{3}{c}{G2: Extended} & \multicolumn{3}{c}{G3: Detailed} & \multicolumn{3}{c}{G4: Complete} & \multicolumn{3}{c}{Overall} \\ \cmidrule(lr){2-4} \cmidrule(lr){5-7} \cmidrule(lr){8-10} \cmidrule(lr){11-13} \cmidrule(lr){14-16}
Method$\downarrow$ & TPR $\uparrow$ & FPR $\downarrow$ & F1 $\uparrow$ & TPR $\uparrow$ & FPR $\downarrow$ & F1 $\uparrow$ & TPR $\uparrow$ & FPR $\downarrow$ & F1 $\uparrow$ & TPR $\uparrow$ & FPR $\downarrow$ & F1 $\uparrow$ & TPR $\uparrow$ & FPR $\downarrow$ & F1 $\uparrow$ \\ \midrule
CLIPScore~\cite{hessel2021clipscore} & 42.12 & 16.24 & 53.26 & 20.03 & 11.53 & 30.41 & 13.65 & 9.18 & 22.23 & 11.53 & 8.94 & 19.14 & 21.82 & 11.47 & 32.74 \\ 
NegCLIPScore~\cite{negclip} & 43.53 & 26.59 & 51.18 & 36.24 & 24.94 & 44.97 & 31.29 & 24.71 & 40.12 & 34.12 & 27.76 & 42.15 & 36.29 & 26.00 & 44.72 \\ 
GPT-4o (DP)~\cite{openai2024gpt4o}
 & 87.61 & \textbf{1.77} & 92.52 & 67.26 & \textbf{3.54} & 78.76 & 53.98 & \textbf{2.65} & 68.93 & 43.36 & \textbf{2.88} & 60.12 & 63.05 & \textbf{2.71} & 76.31 \\ 
GPT-4o (CoT)~\cite{openai2024gpt4o}& 91.15 & 2.65 & 94.06 & 81.42 & 8.85 & 85.58 & 69.91 & 7.08 & 79.08 & 61.95 & 9.73 & 72.16 & 76.11 & 7.08 & 83.09 \\ 
Llama-3.2-90B (DP)~\cite{llama32} & 57.52 & 38.05 & 58.82 & 38.05 & 26.55 & 46.23 & 39.82 & 41.59 & 43.90 & 46.91 & 43.36 & 49.31 & 45.58  & 37.39 & 49.82 \\ 
Llama-3.2-90B (CoT)~\cite{llama32}& 76.99 & 65.49 & 63.50 & 56.64 & 46.02 & 55.90 & 57.52 & 49.56 & 55.55 & 55.75 & 49.56 & 54.31 & 61.73 & 52.65 & 57.59 \\ 
TIFA~\cite{hu2023tifa}  & 80.53 & 53.98 & 68.68 & 78.76 & 30.09 & 75.42 & 75.22 & 56.64 & 64.88 & 76.99 & 67.26 & 63.04 & 77.88 & 51.99 & 67.76  \\ 
VDC~\cite{zhu2023vdc} & 95.33 & 52.05 & 77.09 & 91.33 & 58.04 & 73.26 & 86.00 & 72.67 & 66.49 & 82.07 & 78.09 & 63.08 & 88.67 & 65.17 & 69.86 \\ 
\midrule
HMGIE (Ours) & \textbf{97.65} & 2.35 & \textbf{97.65} & \textbf{96.71} & 8.24 & \textbf{94.37} & \textbf{94.12} & 12.94 & \textbf{90.91} & \textbf{91.29} & 16.24 & \textbf{87.98} & \textbf{94.94} & 9.94 & \textbf{92.68} \\ 
\bottomrule
\end{tabular}%
}
\vspace{-4pt}
\end{table*}
\section{Experiments}
\label{sec:exp}
\subsection{Experimental Setup}
\label{subsec:setup}
\noindent\textbf{Dataset.} We evaluate our approach on the MVTID dataset, which consists of $3,400$ image-caption pairs across four granularities. Beyond data cleansing, we extend our evaluation to two application scenarios: detecting fake news with NewsCLIPpings~\cite{luo2021newsclippings}, a text-image fake news dataset, and TIIL~\cite{huang2024dtiil}, which offers fine-grained fake news data for evaluating text-image mismatches; and assessing text-to-image generation consistency with SeeTRUE~\cite{yarom2024you}, a benchmark focused on measuring alignment accuracy and consistency in generated content.

\noindent\textbf{Implementation Details.} 
Our framework utilizes \texttt{GPT-4o}~\cite{openai2024gpt4o} as the base model for all LLM and MLLM components, with the temperature set to $0.3$. For dataset construction, we employ an ensemble of three MLLMs:  \texttt{GPT-4o},  \texttt{Claude 3.5 Haiku}~\cite{claude}, and  \texttt{Gemini-1.5 Flash}~\cite{team2024gemini} to generate initial captions at each granularity level. \texttt{GPT-4o} is also used for the semantic fusion of generated captions, generation of adversarial perturbations, and evaluation of perturbation detectability. The maximum evaluation level is set to five. Specific prompts designed for each module are given in Appendix~\ref{app:prompt}.
For H-Score calculations, both $\omega$ and $\alpha$ follow a geometric sequence with a ratio of $1.2$.

\noindent\textbf{Baselines.} 
We evaluate HMGIE against several competitive baselines. For embedding-based methods, we compare with CLIPScore~\cite{hessel2021clipscore} and NegCLIPScore~\cite{negclip}, both using a threshold of $0.25$. 
or prompt-based methods, we implement Direct-Prompt (DP) and Chain-of-Thought (CoT)\cite{wei2022chain} approaches using both \texttt{GPT-4o}\cite{openai2024gpt4o} and \texttt{Llama-3.2-90B-Vision}~\cite{llama32} as the backbone models.
For QA-based methods, we compare against TIFA~\cite{hu2023tifa} and VDC~\cite{zhu2023vdc}, two recent approaches that use question answering to detect inconsistencies.

\noindent\textbf{Evaluation Metrics.} For consistency decision evaluation, we use the consistency decision $d^L$ to calculate the following metrics: (1) TPR or recall, denoting the percentage of pairs correctly identified as inconsistent.
(2) FPR, representing the percentage of consistent pairs incorrectly identified as inconsistent.
(3) F1 Score, providing a balanced measure of detection performance.
Additionally, we utilize Kendall's $\tau$~\cite{abdi2007kendall} to assess the correlation between H-Scores and human judgments.

\subsection{Main Results}
\label{subsec:main_results}

\noindent\textbf{Comparison with Existing Methods across Granularities.}
We first \ref{tab:res_main} compares HMGIE with existing methods on the MVTID dataset. 
In general, as shown in Table \ref{tab:res_main}, HMGIE achieves the highest TPR of 94.94\% and F1 score of 92.68\%, with a low FPR of 9.94\%, outperforming the second-best method VDC by 22.82\% in terms of F1 score. This substantial improvement demonstrates the effectiveness of our hierarchical evaluation framework. Moreover, HMGIE exhibits robust performance across different granularity levels, while baseline methods show significant degradation on complex captions. For instance, Direct-Prompt (GPT-4o) experiences a dramatic performance drop of 44.25\% in TPR from G1 to G4. Meanwhile, HMGIE maintains stable performance with only a 6.36\% decrease, validating our approach's adaptability to varying granularities of inconsistency detection.

\begin{table}[t]
\centering
\caption{Kendall’s $\tau$ correlation between semantic consistency metrics and human judgments across different granularities.}
\label{tab:correlation}
\renewcommand\arraystretch{1}
\resizebox{0.75\columnwidth}{!}{%
\begin{tabular}{lccccc}
\toprule
\multirow{2}{*}{Method} & \multicolumn{4}{c}{Granularity} & \multirow{2}{*}{Mixed}\\
\cmidrule(lr){2-5}
& G1 & G2 & G3 & G4 \\
\midrule
CLIPScore~\cite{hessel2021clipscore} & 0.405 & 0.412 & 0.397 & 0.363 &0.203\\
$\mathcal{H}_{acc}$ (Ours) & 0.412 & 0.463 & 0.552 & 0.598 &0.640\\
\bottomrule
\end{tabular}}
\end{table}

\begin{table}[t]
\centering
\caption{Kendall’s $\tau$ correlation between semantic completeness metrics and human judgments across different granularities.}
\label{tab:correlation_comp}
\renewcommand\arraystretch{1}
\resizebox{0.75\columnwidth}{!}{%
\begin{tabular}{lccccc}
\toprule
\multirow{2}{*}{Method} & \multicolumn{4}{c}{Granularity}&\multirow{2}{*}{Mixed} \\
\cmidrule(lr){2-5}
& G1 & G2 & G3 & G4 \\
\midrule
Text Length & 0.594 & 0.520 & 0.501 & 0.511&0.590 \\
$\mathcal{H}_{comp}$ (Ours) & 0.656 & 0.621 & 0.614 & 0.697 &0.683 \\
\bottomrule
\end{tabular}}
\vspace{-3pt}
\end{table}

\noindent\textbf{H-Scores Correlation with Human Judgment.}
Following prior works~\cite{hessel2021clipscore,hu2023tifa,Cho2024DSG}, we evaluate the correlation between our proposed H-Scores and human judgments. We randomly sample 400 image-caption pairs (100 from each granularity level) and collect annotations from 10 human annotators who rate semantic consistency and semantic completeness using a five-point Likert scale (1: strongly disagree to 5: strongly agree). For consistency rating, annotators assess whether the caption accurately matches the image content. For completeness rating, they evaluate whether the caption covers all significant visual elements. Table \ref{tab:correlation} and \ref{tab:correlation_comp} present Kendall's $\tau$ correlation coefficients between different metrics and human Likert ratings across different granularity levels. 
As shown in Table \ref{tab:correlation}, $\mathcal{H}{acc}$ consistently demonstrates a stronger correlation with human judgments compared to CLIPScore \cite{hessel2021clipscore}, particularly with substantial improvements in detailed descriptions (G3, G4). This is primarily because CLIPScore relies on global image-text embeddings and cannot distinguish fine-grained semantic elements. The performance gap becomes more pronounced on mixed granularity data, as CLIPScore struggles to assess consistency across different granularity levels. This validates that HMGIE's hierarchical questioning and evaluation process better aligns with how humans evaluate semantic consistency. 
For semantic completeness assessment, we compare $\mathcal{H}_{comp}$ with a simple baseline that directly uses normalized text length as a completeness score. 
Table~\ref{tab:correlation_comp} shows that $\mathcal{H}_{comp}$ significantly outperforms this baseline based on length at all granularity levels, indicating that our graph-based semantic coverage tracking provides a more reliable assessment of description completeness compared to surface-level text statistics.

\noindent\textbf{Comparison of Extended Applications.}
We extend HMGIE’s capability beyond visual-textual data cleansing to its practical utility in fake news detection and text-to-image generation evaluation. Table \ref{tab:prac_app} presents the evaluation results. Specifically, on the fake news datasets NewsCLIPpings~\cite{luo2021newsclippings} and TIIL~\cite{huang2024dtiil}, HMGIE achieves TPRs of 95.56\% and 96.46\%, respectively, demonstrating strong efficacy in identifying manipulated content. Additionally, HMGIE outperforms other methods on the SeeTRUE~\cite{yarom2024you} dataset, which assesses text-to-image consistency, with a TPR of 76.99\%. Through experiments, the observed relatively high FPR may be attributed to data quality issues, as some genuine samples contain subtle inconsistencies, and also because validating news often requires out-of-context information. Examples can be found in Appendix~\ref{app: fpr_fail_result}.

\begin{table}[t]
\centering
\caption{Performance comparison (TPR in \%) of fake news detection and text to image evaluation  across different datasets.}
\label{tab:prac_app}
\renewcommand\arraystretch{1}
\resizebox{\linewidth}{!}{%
\begin{tabular}{@{}lccc@{}}
\toprule
Method$\downarrow$ & NewsCLIPpings~\cite{luo2021newsclippings} & TIIL~\cite{huang2024dtiil} & SeeTRUE~\cite{yarom2024you} \\ \midrule
CLIPScore~\cite{hessel2021clipscore} & 23.01 & 87.61  & 48.67 \\
NegCLIPScore~\cite{negclip} & 68.41 & 85.84  & 37.17 \\ 
GPT-4o (DP)~\cite{openai2024gpt4o} & 47.79 & 89.38  & 63.72 \\ 
GPT-4o (CoT)~\cite{openai2024gpt4o} & 31.86 & 92.04  & 70.80 \\ 
Llama-3.2-90B (DP)~\cite{llama32}& 62.83 & 86.61  & 53.98 \\ 
Llama-3.2-90B (CoT)~\cite{llama32}& 86.73 & 89.29  & 69.65 \\ 
TIFA~\cite{hu2023tifa} & 91.15 & 87.58  & 72.30 \\ 
VDC~\cite{zhu2023vdc} & 71.68 & 92.92  & 57.52 \\ 
\midrule
HMGIE (Ours) & \textbf{95.56} & \textbf{96.46}  & \textbf{76.99} \\ 
\bottomrule
\end{tabular}%
}
\end{table}

\subsection{Ablation Studies}
\label{subsec:ablation}
\begin{figure}[t]
    \centering
    \includegraphics[width=\linewidth]{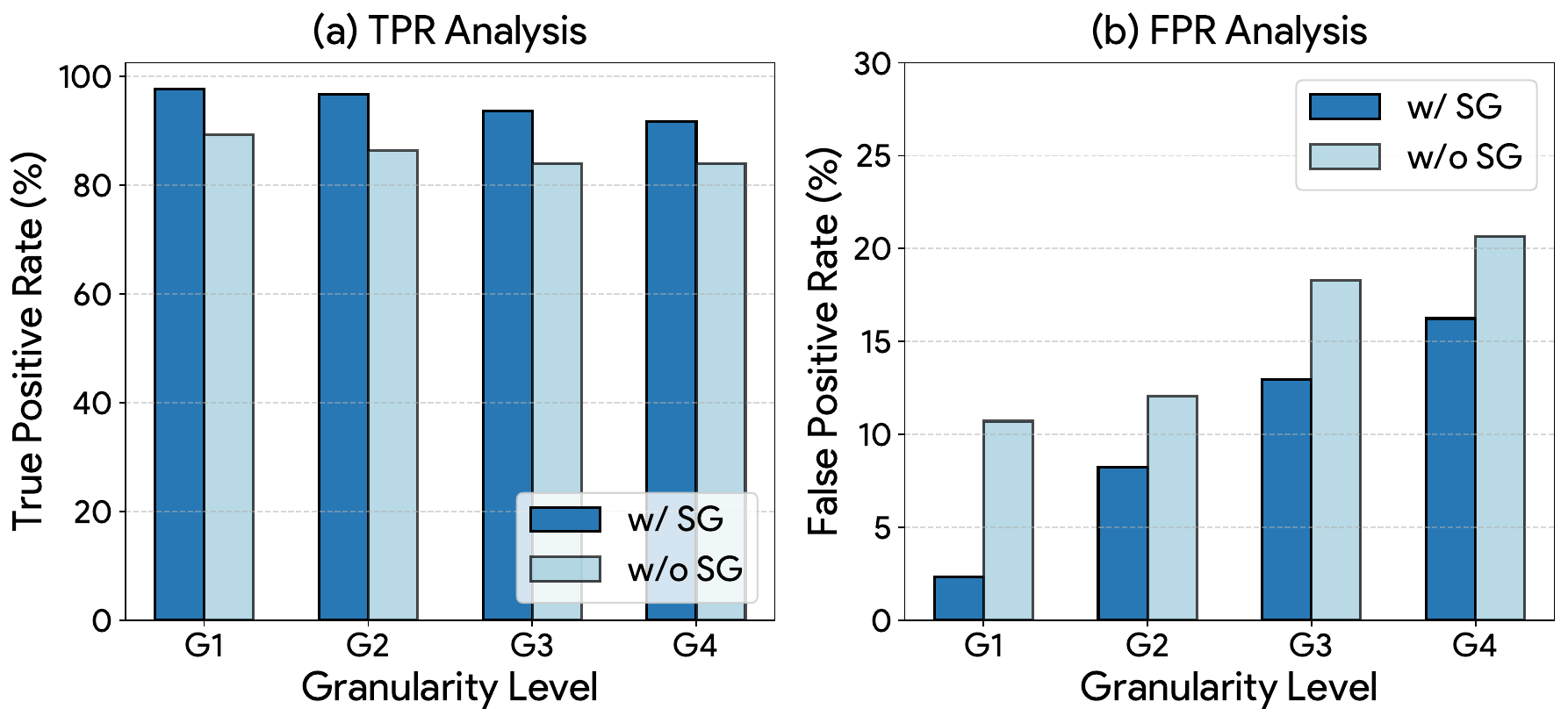}
    \vspace{-7pt}
    \caption{Ablation study on semantic graph (SG) in HMGIE: (a) TPR and (b) FPR comparison across granularity levels.}
    
    \label{fig:ablation}
\end{figure}

\noindent\textbf{Impact of Semantic Graph.}
To validate the effectiveness of semantic graph parsing in HMGIE, we conducted ablation studies by replacing the semantic graph with textual captions for question generation and evaluation. As shown in Figure~\ref{fig:ablation}, removing semantic graph parsing consistently degrades performance across all granularity levels. Specifically, the true positive rate (TPR) drops by an average of 9.01\%, with the largest decrease observed in G2. In addition, false positive rate (FPR) results indicate a significant increase, with an average FPR rise of 5\% when the semantic graph is removed. These findings confirm that semantic graph parsing is essential for maintaining high true detection rates and minimizing false alarms, thereby validating its crucial role in achieving precise semantic understanding across varying levels of granularity.

\begin{figure}[t]
	\centering
	\begin{minipage}[c]{0.44\linewidth}
		\centering
		\includegraphics[width=\textwidth]{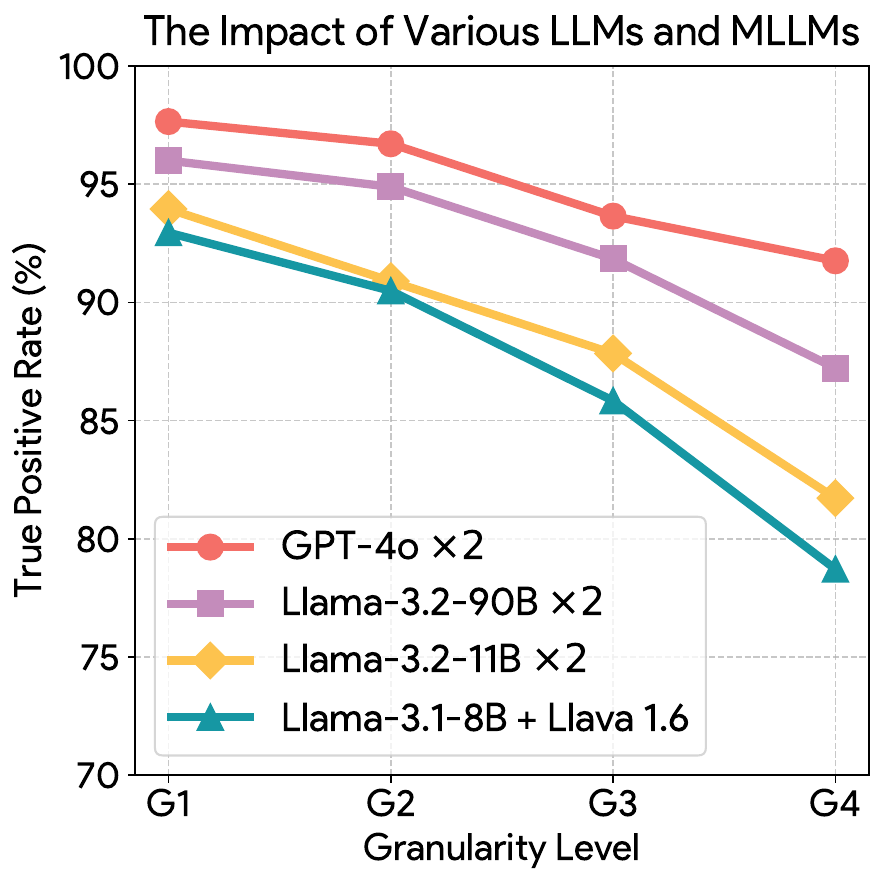}
		\caption{Impact of different LLMs and MLLMs.}
        \vspace{-10pt}
		\label{fig:llm}
	\end{minipage} 
        \hspace{0.02\textwidth}
	\begin{minipage}[c]{0.44\linewidth}
		\centering
		\includegraphics[width=\textwidth]{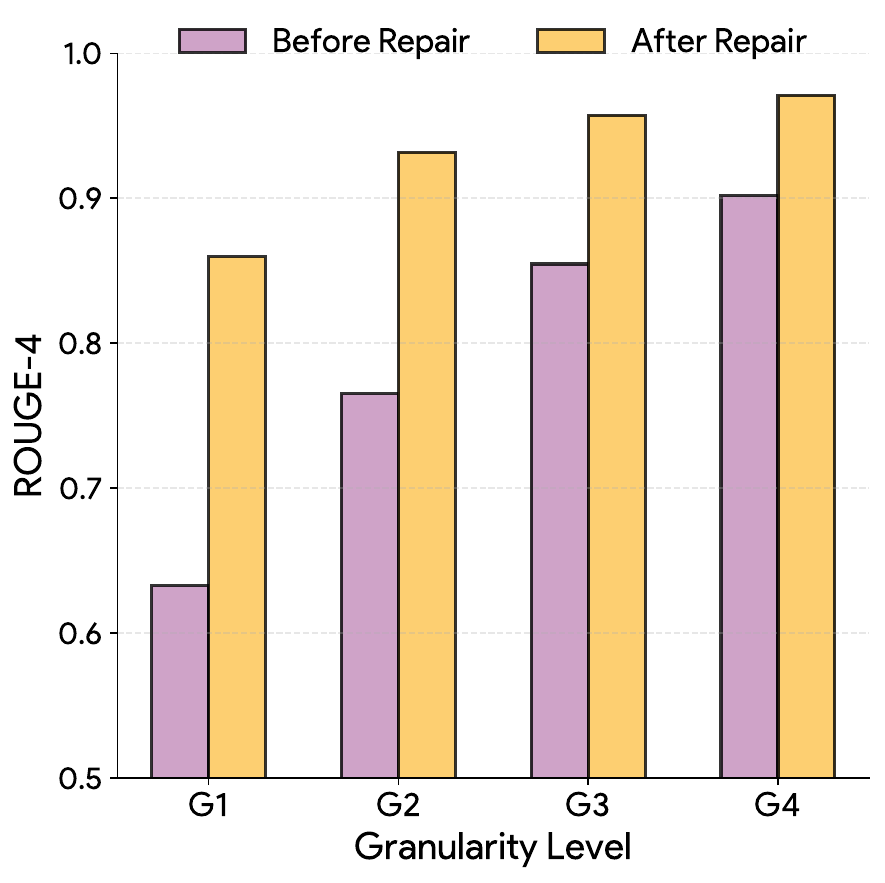}
		\caption{ROUGE-4 for captions before and after repair.}
        \vspace{-10pt}
		\label{fig:repaire}
	\end{minipage}
\end{figure}

\noindent\textbf
{Impact of Various LLMs and MLLMs.}
To evaluate the flexibility and robustness of our framework, we conducted experiments by replacing the \texttt{GPT-4o} in HMGIE with various open-source alternatives, using \texttt{Llama-3.2-90B}~\cite{llama32}, and \texttt{Llama-3.2-11B}~\cite{llama32} as both LLM and MLLM components, as well as a combination of \texttt{Llama-3.1-8B}~\cite{dubey2024llama} and \texttt{Llava-1.6}~\cite{llava} for LLM and MLLM respectively. 
As shown in Figure~\ref{fig:llm}, GPT-4o achieves the best performance with TPR across all granularity.
While there is a performance gap between open-source and default proprietary models, the difference is relatively small, \eg \texttt{Llama-3.2-90B} demonstrates strong performance with TPR ranging from 96\% to 86\%.
We find that larger model size are more suitable for inconsistency detection, as \texttt{Llama-3.2-90B} consistently outperforms \texttt{Llama-3.2-11B} by 2-4\% TPR.
Notably, the performance gaps between different models become more pronounced as granularity increases, indicating that more capable models are particularly advantageous for handling complex, detailed descriptions.

\subsection{Qualitative Analysis}
\label{subsec:qualitative}
\begin{figure}[t]
    \centering
    \includegraphics[width=\linewidth]{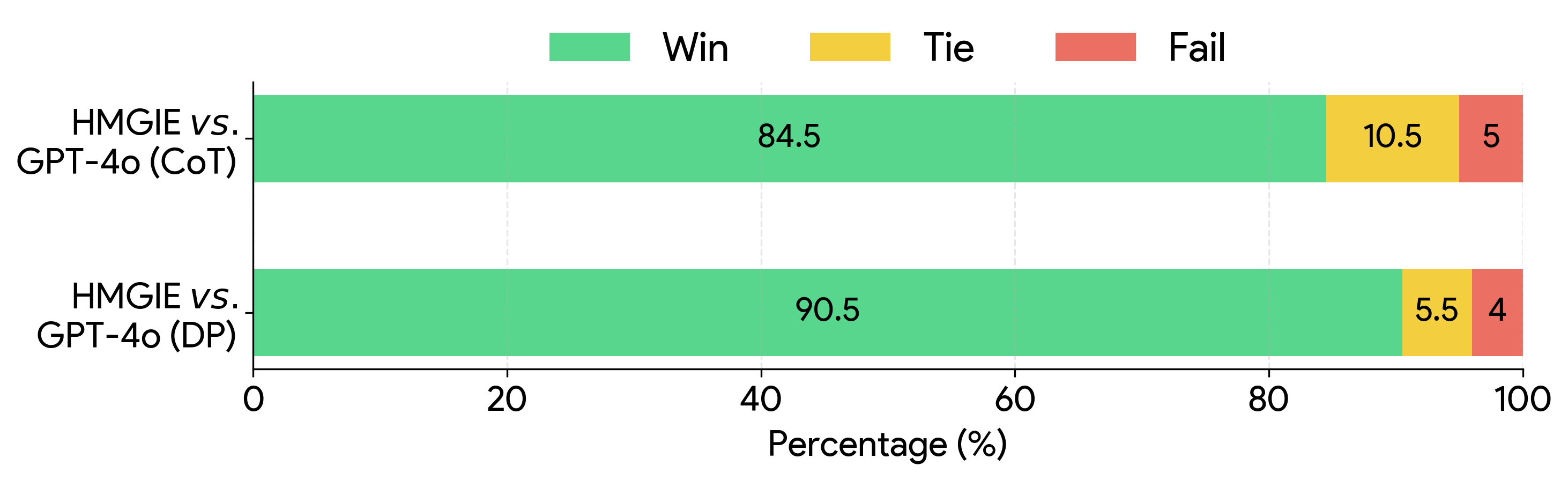}
    \caption{Comparison of explanation quality between HMGIE and baseline methods.}
    \vspace{-10pt}
    \label{fig:explain_evaluate}
\end{figure}

\noindent\textbf{Interpretability Evaluation.}
To assess the explanatory quality of different methods, we conduct a comparative evaluation between HMGIE's explanations and those generated by Direct-Prompt and CoT-Prompt methods with 
\texttt{GPT-4o}. We present human annotators with inconsistency cases accompanied by explanations from each method, asking them to select the most accurate and comprehensible explanation. The results shown in Figure \ref{fig:explain_evaluate} demonstrate HMGIE's clear superiority in providing interpretable explanations. When compared against CoT-Prompt, HMGIE's explanations were preferred in 84.5\% of cases and only 5\% cases favoring CoT-Prompt. The advantage is even more pronounced when compared to Direct-Prompt, with HMGIE winning 90.5\% of comparisons. This substantial performance gap can be attributed to HMGIE's hierarchical evaluation framework, which constructs a clear reasoning path through the HIEG, enabling more structured and comprehensive explanations of detected inconsistencies.

\noindent\textbf{Caption Repair Evaluation.}
We further evaluate HMGIE's caption repair capability by comparing ROUGE~\cite{lin2004rouge} scores between the captions before and after repair against the original ground-truth captions. The results in Figure \ref{fig:repaire} show that after repair, the captions achieve significantly higher ROUGE-4 scores ranging from 0.85 to 0.97 across all granularities, while the scores before repair only range from 0.68 to 0.82, demonstrating that HMGIE can effectively identify and correct inconsistencies while preserving the original semantic content.

\section{Conclusion}
\label{sec:conclusion}
This paper proposes HMGIE, a hierarchical and multi-grained framework for visual-textual inconsistency evaluation. Our method addresses the challenges of various caption granularities and diverse inconsistency types through three key components: semantic graph generation for structured understanding, hierarchical inconsistency evaluation for progressive evaluation, and quantitative evaluation for comprehensive assessment. Through extensive experiments on our proposed MVTID dataset and other benchmarks, HMGIE demonstrates superior performance in evaluating multi-grained inconsistencies, achieving significant improvements over existing methods.

\noindent\textbf{Discussion.} HMGIE effectively leverages the advanced reasoning capabilities of current LLMs and benefits from their continued evolution.  We acknowledge that its efficiency is constrained by the computational overhead of these models. However, this limitation can be addressed through recent advancements~\cite{xiao2023smoothquant,shen2021efficient,dao2022flashattention} in model efficiency optimization. Furthermore, our modular design allows for the replacement of general large models with more efficient or task-specific models (see Figure ~\ref{fig:llm}). This flexibility ensures that HMGIE can adapt to various computational constraints while maintaining its visual-textual inconsistency evaluation effectiveness.

{
    \small
    \bibliographystyle{ieeenat_fullname}
    \bibliography{main}
}

\clearpage
\appendix
\setcounter{page}{1}
\maketitlesupplementary

\section*{Outline of Appendix}
\begin{itemize}
    \item MVTID Dataset Details (\ref{app:dataset})
    
    \item Method Details (\ref{app:method})
        \begin{itemize}
            \item Prompts used in LLMs and MLLMs (\ref{app:prompt})
            \item Algorithm Details (\ref{app:alg})
        \end{itemize}
    
    \item Additional Experimental Results (\ref{app:experiment})
        \begin{itemize}
            \item Visualization of Evaluation Process of HMGIE (\ref{app: visual})
                \begin{itemize}
                    \item Examples of Semantic Graph (\ref{app: sg})
                    \item Examples of HIEG (\ref{app: HIEG})
                \end{itemize}
            \item Failure Case Analysis in Extended Benchmark Datasets (\ref{app: fpr_fail_result})
                \begin{itemize}
                    \item NewsCLIPpings (\ref{app: fpr_fail_result_newsclipings})
                    \item TIIL (\ref{app: fpr_fail_result_tiil})
                    \item SeeTRUE (\ref{app: fpr_fail_result_seetrue})
                \end{itemize}
            \item Impact of HMGIE-based Data Cleansing on Vision-Language Model Performance (\ref{app: blip_finetuning})
        \end{itemize}
\end{itemize}  

\section{MVTID Dataset Details}
\label{app:dataset}
\begin{figure*}
    \centering
    \includegraphics[width=\linewidth]{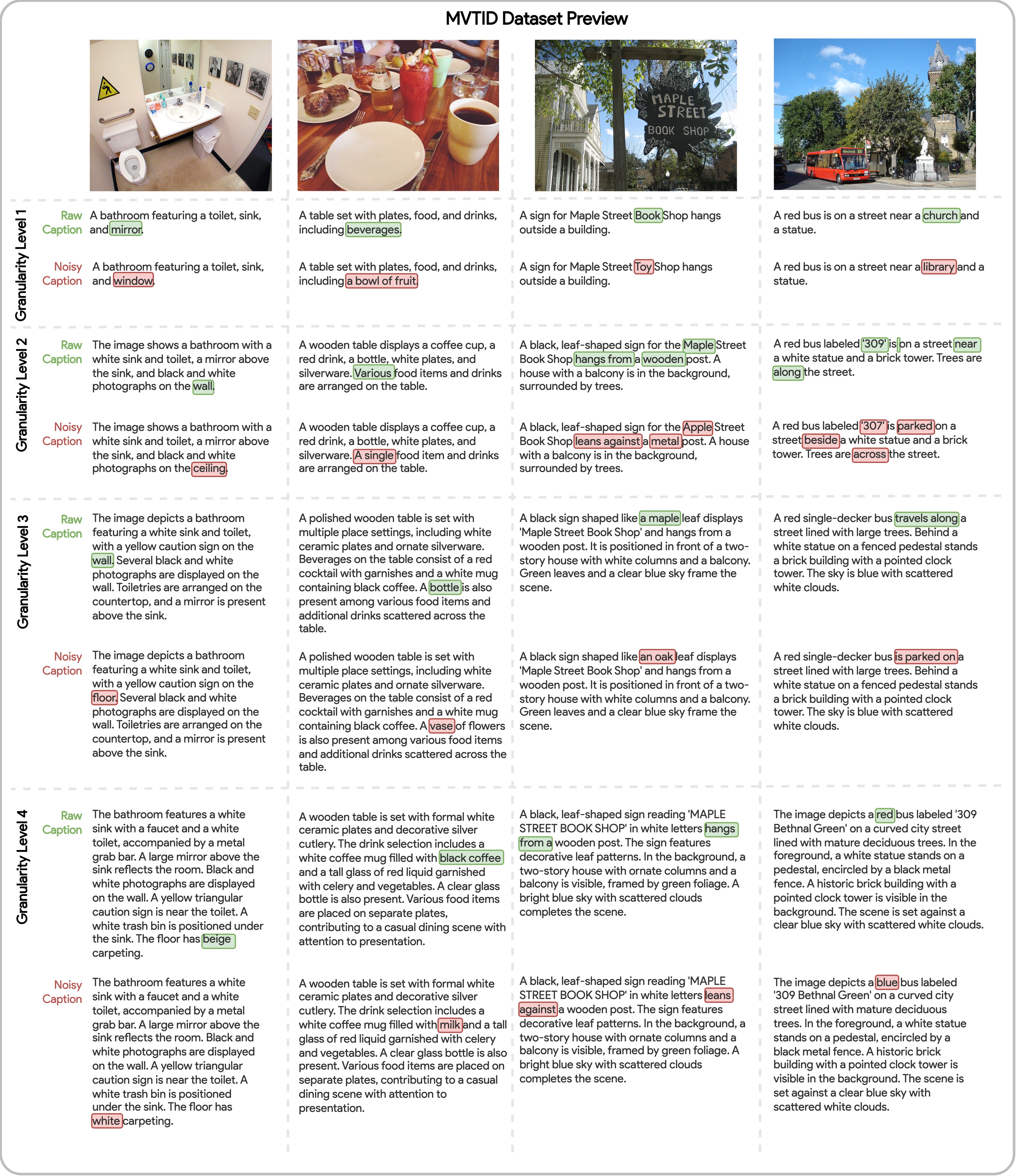}
    \caption{Examples of MVTID dataset.}
    \label{fig:dataset}
\end{figure*}

The Multi-granularity Visual-Textual Inconsistency Dataset (MVTID) consists of carefully curated image-caption pairs designed to evaluate visual-textual inconsistency detection across varying levels of descriptive granularity. As illustrated in Figure ~\ref{fig:dataset}, we present representative examples that demonstrate the dataset's key characteristics and the progression of caption complexity across four distinct granularity levels. Taking the first bathroom scene as an example:

At G1 (Basic), the caption focuses solely on core objects in the scene (toilet, sink, mirror). The inconsistency is introduced through a simple object substitution (``mirror" → ``window"), representing the most basic form of semantic error.

At G2 (Extended), the description incorporates basic visual attributes (white sink, black and white photographs) and spatial relationships (above the sink, on the wall). The inconsistency shifts to a spatial relation error by modifying the location of the photographs (``wall" → ``ceiling"), while maintaining the same descriptive elements.

At G3 (Detailed), the caption expands to include secondary objects (yellow caution sign, toiletries) and more precise spatial configurations. The inconsistency becomes more sophisticated, involving the spatial relationship of a prominent feature (caution sign's location from ``wall" → ``floor"), which requires understanding both object placement and scene layout.

At G4 (Complete), the description encompasses all visual elements with precise attributes and comprehensive spatial relationships. The inconsistency appears in fine-grained attribute details (carpet color: ``beige" → ``white"), demonstrating how subtle changes in specific object properties can create semantic misalignments even within detailed scene descriptions.

\section{Method Details}
\label{app:method}

\subsection{Prompts Used in LLMs and MLLMs}
\label{app:prompt}
Table~\ref{tab:prompt_dp} shows the prompt template used for direct prompting (DP) evaluation, which is implemented as one of our baseline methods. This prompt directly asks the model to determine if an image and caption pair is consistent.

Table~\ref{tab:prompt_cot} shows the prompt template used for chain-of-thought (CoT) evaluation, another baseline method that guides the model through step-by-step reasoning about image-caption consistency.

Table~\ref{tab:prompt_sg} shows the prompt template used for semantic graph generation. This prompt guides the LLM to convert caption text into a structured semantic graph representing entities and their relationships.

Table~\ref{tab:prompt_qg} shows the prompt template used for hierarchical question generation. Based on the semantic graph and evaluation level, this prompt helps generate targeted questions to evaluate specific semantic elements.

Table~\ref{tab:promt_vqa} shows the prompt template used for visual question answering. This prompt instructs the MLLM to answer the generated questions by analyzing the image content.

Table~\ref{tab:promt_qe} shows the prompt template used for question-answer evaluation. This prompt helps determine whether the VQA-generated answers are semantically equivalent to the reference answers.

Table~\ref{tab:promt_verify} shows the prompt template used for semantic coverage analysis. This prompt analyzes which semantic elements in the semantic graph have been verified, and provides suggestions for question generation in the next evaluation level to ensure comprehensive coverage.

Table~\ref{tab:promt_explain} shows the prompt template used for natural language explanation generation. This prompt synthesizes the evaluation results from HIEG to produce clear and informative explanations about the detected inconsistencies or confirmations of consistency.

\subsection{Algorithm Details}
\label{app:alg}
Algorithm~\ref{alg:hmgie} presents the complete pseudocode of our Hierarchical and Multi-Grained Inconsistency Evaluation (HMGIE) framework. The algorithm takes an image and its corresponding caption as input, and outputs the H-Scores, consistency decision, along with natural language explanation. The pseudocode illustrates how the three main modules of HMGIE work together to achieve comprehensive visual-textual inconsistency evaluation.

\section{Additional Experimental Results}
\label{app:experiment}

\subsection{Visualization of Evaluation Process of HMGIE}
\label{app: visual}
To provide better understanding of HMGIE's internal evaluation mechanism, we present visualizations of two components in our framework: the semantic graph construction and the hierarchical inconsistency evaluation process.

\subsubsection{Examples of Semantic Graph}
\label{app: sg}
\begin{figure*}[!h]
    \centering
    \includegraphics[width=0.85\linewidth]{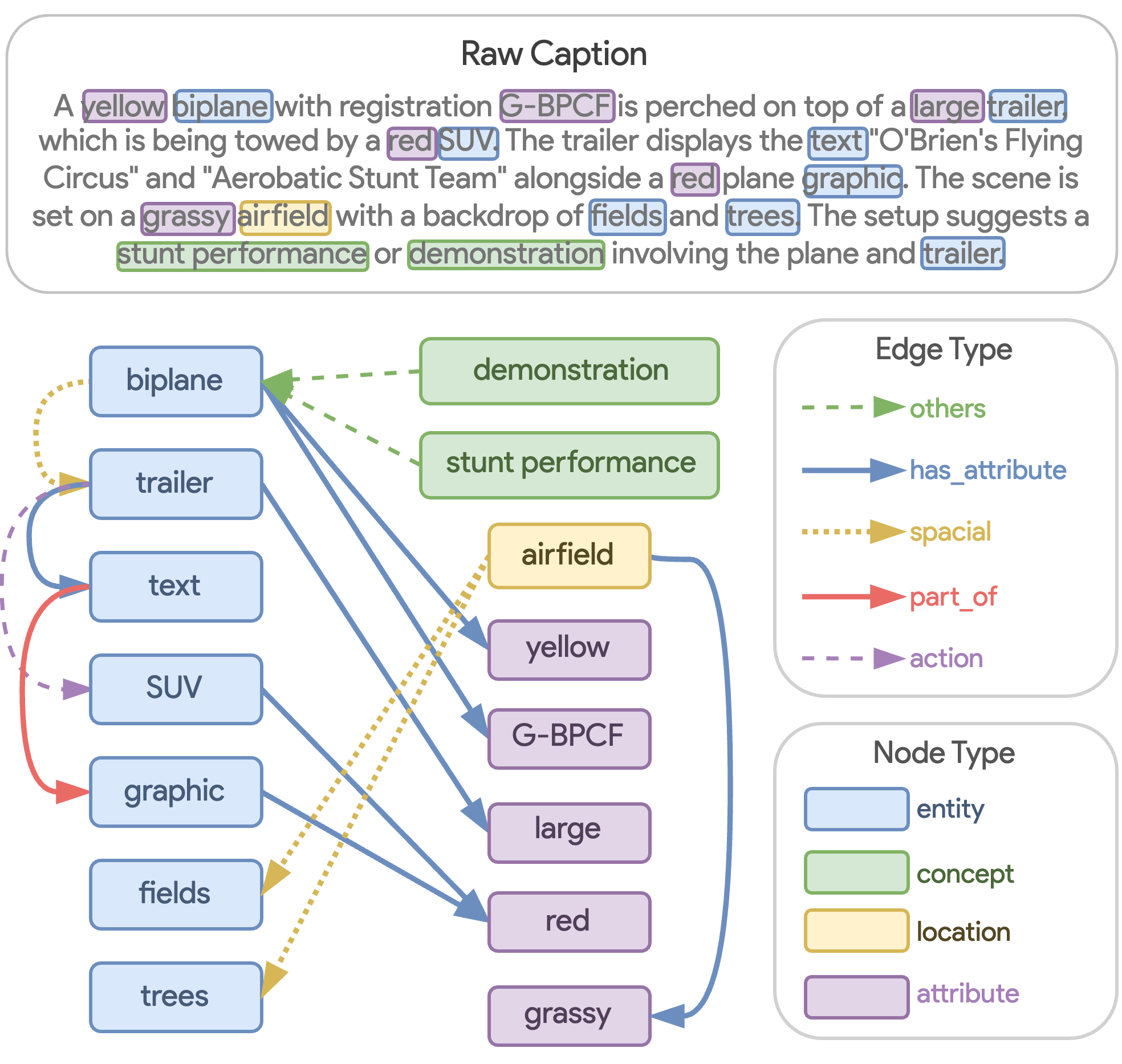}
    \caption{Example of semantic graph for raw caption.}
    \label{fig:sg1}
\end{figure*}

\begin{figure*}[!h]
    \centering
    \includegraphics[width=0.85\linewidth]{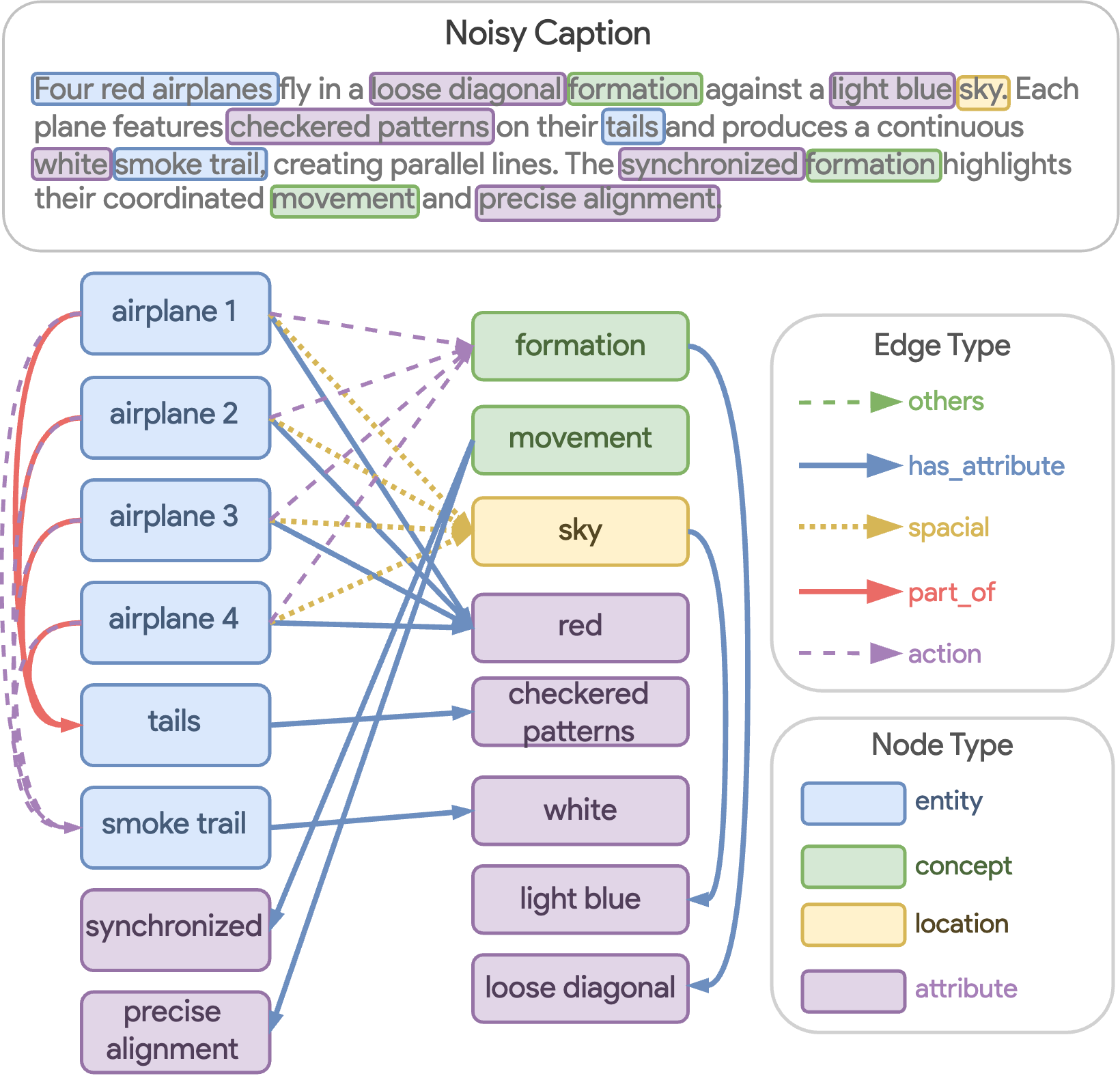}
    \caption{Example of semantic graph for noisy caption.}
    \label{fig:sg2}
\end{figure*}

Figures \ref{fig:sg1} and \ref{fig:sg2} present examples of semantic graph constructed by our framework, where nodes with different colors capture entities, concepts, locations,  attributes and events. These nodes are connected by various types of edges representing relationships such as attributes, spatial relationships), part-of relationships, and others. This structured representation enables systematic analysis and tracking of semantic elements during the evaluation process.

\subsubsection{Examples of HIEG}
\label{app: HIEG}
\begin{figure*}[!h]
    \centering
    \includegraphics[width=0.85\linewidth]{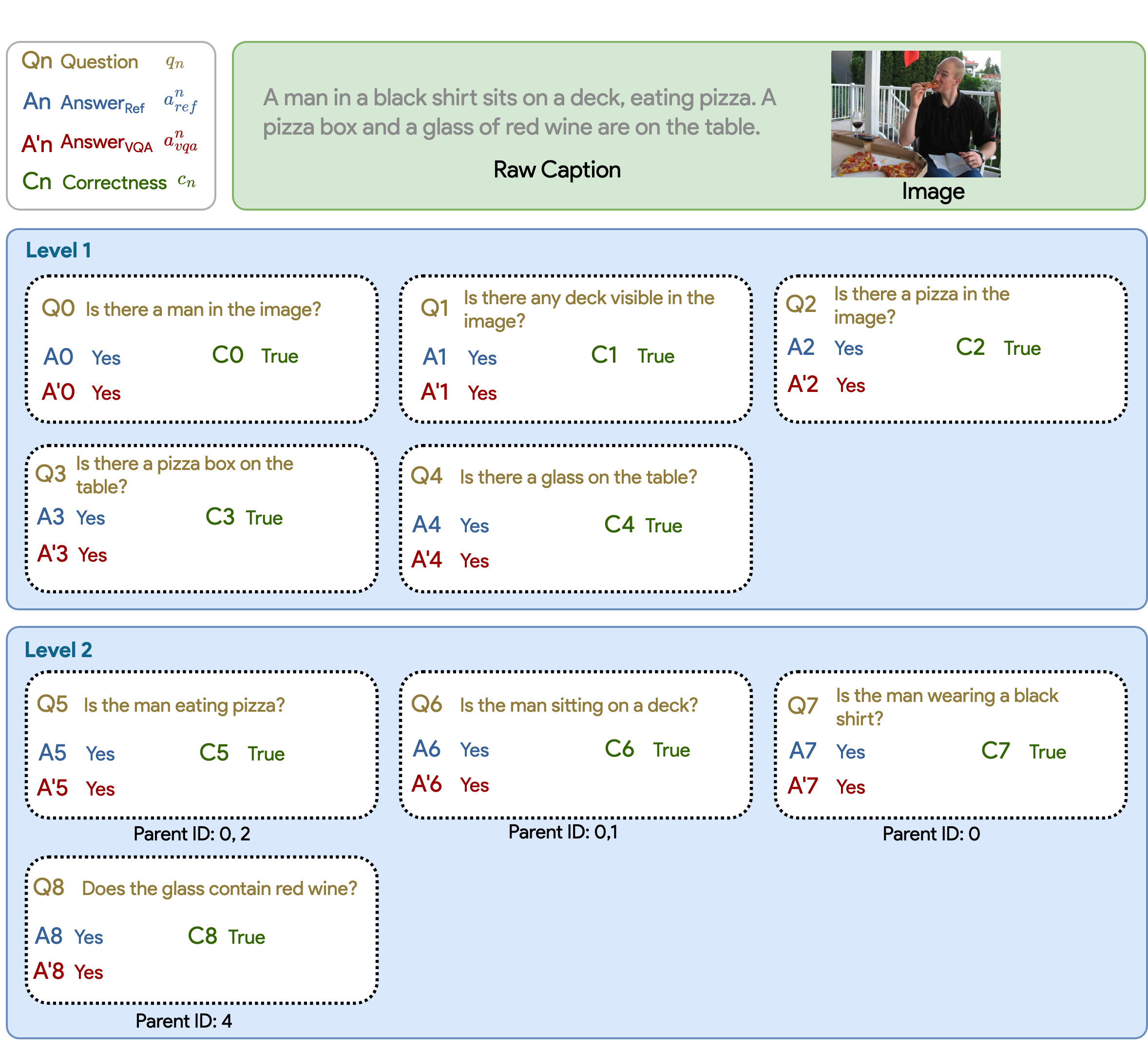}
    \caption{Example of HIEG for the raw caption.}
    \label{fig:hieg1}
\end{figure*}

\begin{figure*}[!h]
    \centering
    \includegraphics[width=0.85\linewidth]{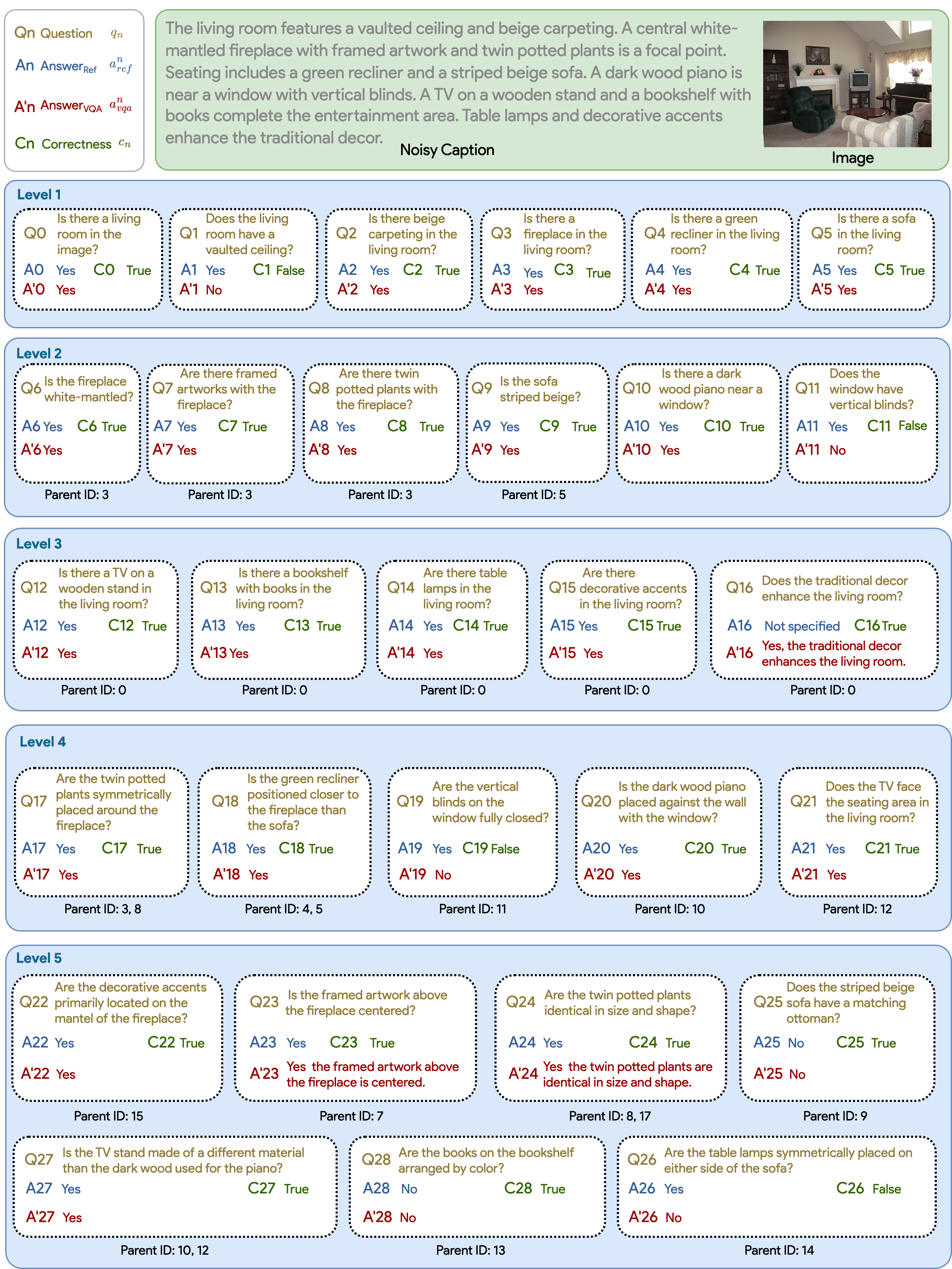}
    \caption{Example of HIEG for the noisy caption.}
    \label{fig:hieg2}
\end{figure*}

Figures \ref{fig:hieg1} and \ref{fig:hieg2} illustrates examples of the Hierarchical Inconsistency Evaluation Graph (HIEG) constructed by our framework. The HIEG demonstrates how evaluations progress from basic semantic elements to more complex relationships across multiple levels, with each node representing a question-answer pair. For clarity of visualization, instead of explicitly drawing edges between nodes, we use parent ID to indicate the hierarchical relationships between evaluation nodes.

\subsection{Failure Case Analysis in Extended Benchmark Datasets}
\label{app: fpr_fail_result}
As discussed in Section \ref{subsec:main_results}, we extended HMGIE's application beyond visual-textual data cleansing to fake news detection (NewsCLIPpings~\cite{luo2021newsclippings}, TIIL~\cite{huang2024exposing}) and text-to-image generation evaluation (SeeTRUE~\cite{yarom2024you}). While HMGIE demonstrates strong performance in detecting inconsistent samples, we also discovered its capability to identify samples that were originally labeled as consistent but actually contain semantic inconsistencies, revealing inherent limitations in existing benchmark datasets. Below, we analyze these cases for each dataset.

\subsubsection{NewsCLIPpings}
\label{app: fpr_fail_result_newsclipings}
For NewsCLIPpings, we present three cases in Table \ref{tab:fpr_examples_newsclip}:
(2)  The second example displays a financial chart with trend lines, but the specific caption about ``shares plummeting in Warsaw" and ``Swiss franc soaring against Zloty" cannot be determined from the generic market visualization alone.
(3) The third example shows a worker at a gas facility, but the caption about ``Ukraine's gas imports" and ``storage levels" cannot be verified from this single operational scene.

While the news themselves might be factual, HMGIE identifies them as potential inconsistencies due to the semantic inconsistencies between the visual content and textual captions. This strict detection strategy, though potentially leading to higher false positive rates, reflects HMGIE's emphasis on verifiable visual-textual alignment in fake news detection.

\subsubsection{TIIL}
\label{app: fpr_fail_result_tiil}
Table \ref{tab:fpr_examples_tiil} shows examples in TIIL. (1) In the first example, while the image shows two artistic sketches, the caption contains details about their weight (80 pounds) and creation time (four months) - information that cannot be determined from visual content alone.
(2) The second example shows flowers in front of a house at dusk, but verifying whether this is specifically ``Barry's home" is impossible without additional context about the location and ownership.
(3) The third example captures a basic scene of a shuttle being transported on a aircraft, but the caption includes extensive details about the date (April 17), location (Dulles airport), and destination (Smithsonian) that go far beyond what is visually apparent.

These cases also reveals limitations in handling claims requiring external knowledge verification, suggesting the need to integrate fact-checking mechanisms for more comprehensive fake news detection.

\subsubsection{SeeTRUE}
\label{app: fpr_fail_result_seetrue}
Table \ref{tab:fpr_examples_seetrue} shows several questionable annotations that were originally labeled as consistent in SeeTRUE dataset: (1) In the first example, the caption states ``many big beds" in the bedroom, while the image shows a simple bedroom setting with clearly only one bed visible. The use of ``many" directly contradicts what can be observed in the image.
(2) The second example shows two scarecrow-like figures in a field, but the caption describes ``a bird scaring a scarecrow".  There is no bird present in the scene at all.
(3) The third example shows two dogs sitting together on grass - one appears to be a tan and white dog and the other a white dog. The caption claims ``one cat and two dogs," but there is no cat visible in the image, only the two dogs.

These cases suggest that many detected inconsistencies are not false positives but rather genuine semantic misalignments that were incorrectly labeled as consistent in the original dataset, highlighting the need for more rigorous quality control measures during dataset creation.

\subsection{Impact of HMGIE-based Data Cleansing on Vision-Language Model Performance}
\label{app: blip_finetuning}

To demonstrate the practical value of our HMGIE framework in improving vision-language models through data cleansing, we conduct comprehensive experiments to evaluate its impact on image captioning performance. Specifically, we investigate how data quality affects model performance by comparing different fine-tuning strategies.

\noindent\textbf{Experimental Setup.} 
We utilize the BLIP-image-captioning-base model from {Huggingface}\footnote{\url{https://huggingface.co/Salesforce/blip-image-captioning-base}} as our baseline architecture. For fine-tuning data, we prepare two versions of the MVTID dataset: (1) the original mixed dataset containing both clean and noisy samples, and (2) a cleansed version processed by our HMGIE framework. We evaluate three model variants:
\begin{itemize}
    \item \textbf{w/o fine-tuning}: The original BLIP-image-captioning-base model without further fine-tuning.
    \item \textbf{FT w/ mixed data}: The model fine-tuned with the original MVTID dataset.
    \item \textbf{FT w/ cleaned data}: The model fine-tuned with the HMGIE-cleaned MVTID dataset.
\end{itemize}
For evaluation, we randomly sample 1,000 images from the Flickr8k\footnote{\url{https://huggingface.co/datasets/atasoglu/flickr8k-dataset}} dataset as our test set. We collect captions generated by all three models and employ GPT-4o as an automated evaluator to determine the best caption among them, ensuring consistent and comprehensive evaluation criteria.

\begin{figure*}[!h]
    \centering
    \includegraphics[width=0.7\linewidth]{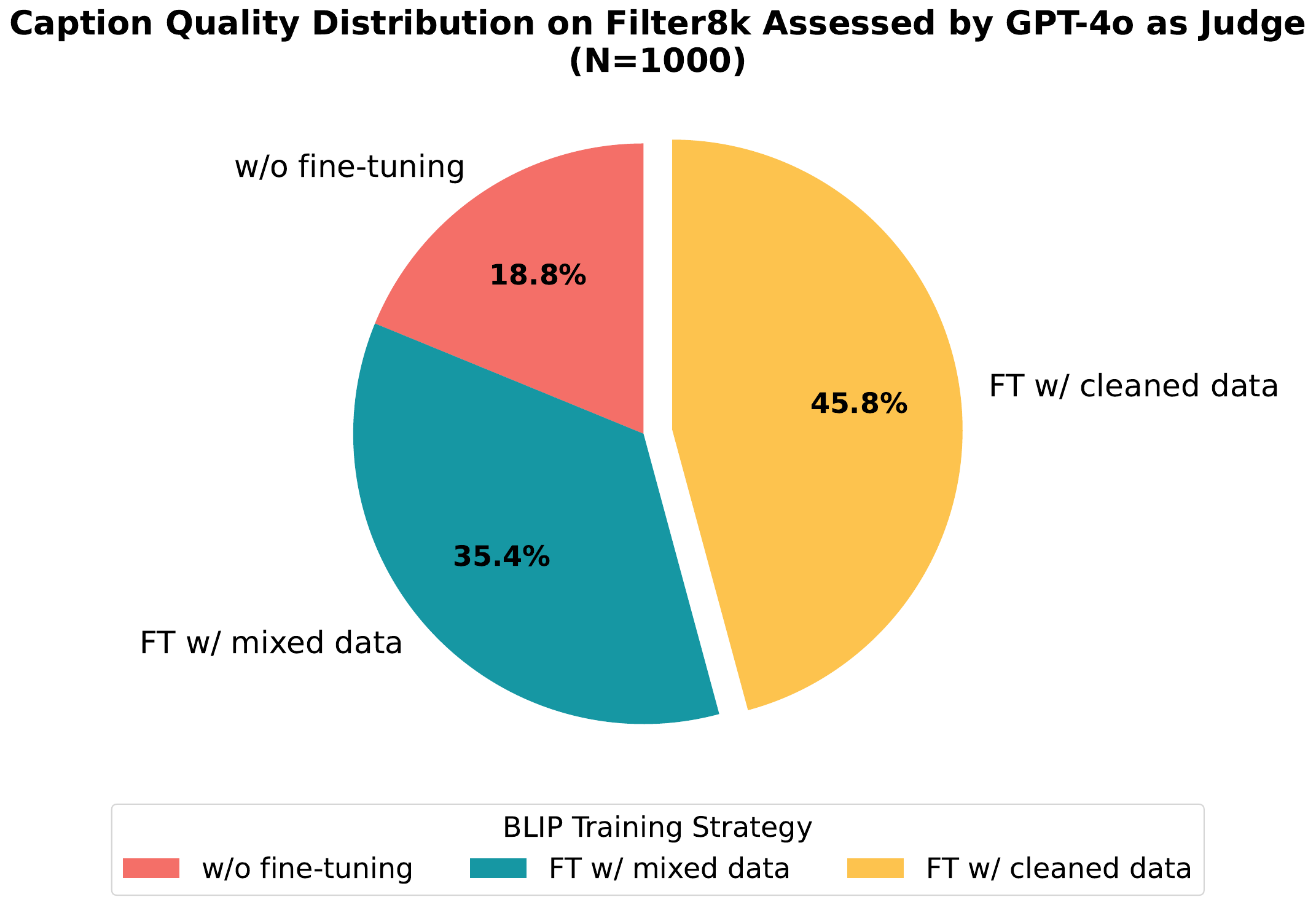}
    \caption{Distribution of caption quality among three models on Flickr8k test set, as evaluated by GPT-4o.}
    \label{fig:blip_finetune}
\end{figure*}

\begin{figure*}
    \centering
    \includegraphics[width=\linewidth]{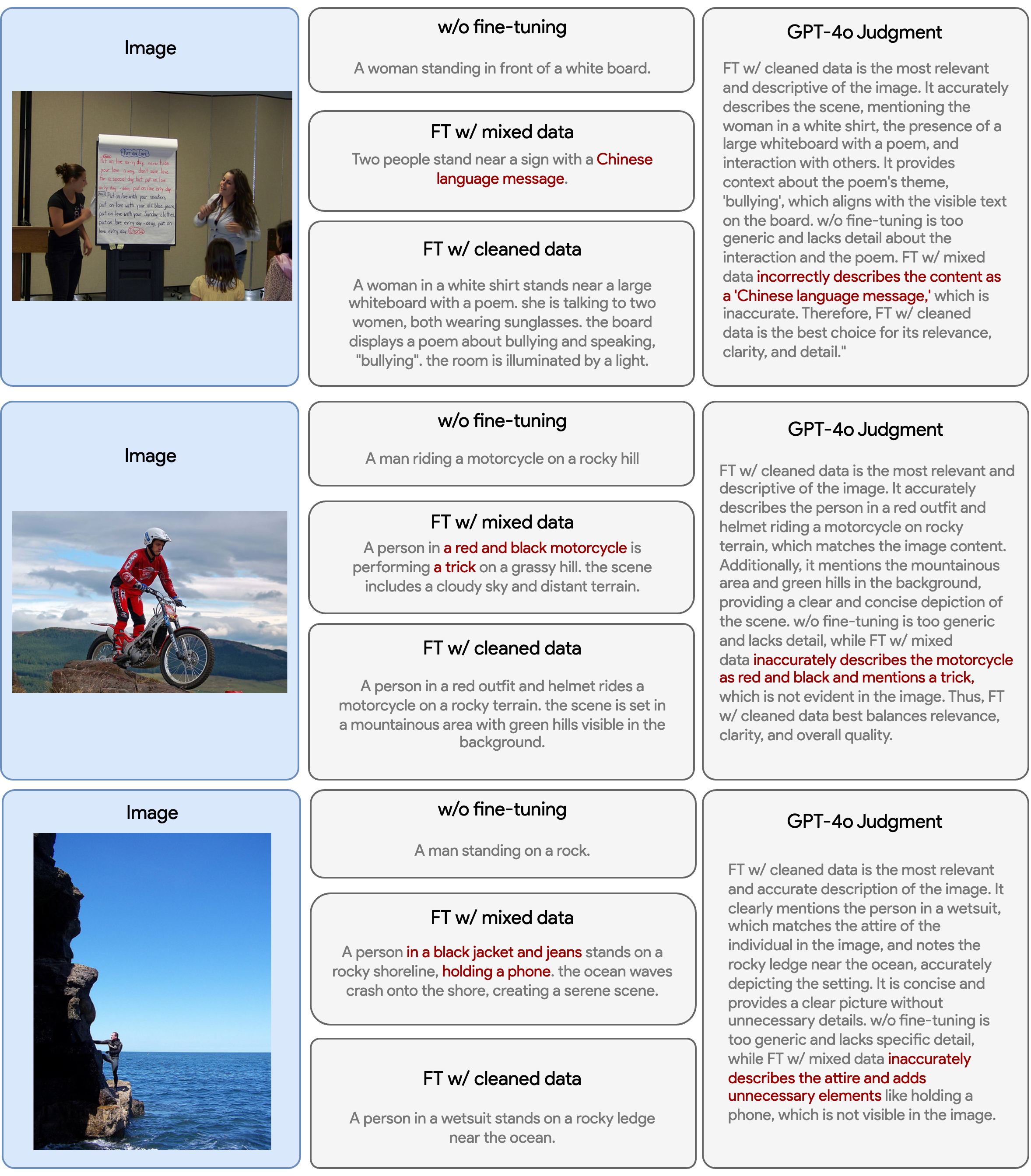}
    \caption{Examples of generated captions and evaluations by {GPT-4o}. Each example consists of an image, three captions generated by different models, and {GPT-4o}'s judgment and explanation of the best caption.}
    \label{fig:table15}
\end{figure*}

\noindent\textbf{Results and Analysis.} 
Figure \ref{fig:blip_finetune} presents the distribution of caption quality scores across the three models. FT w/ cleaned data demonstrates superior performance, generating the highest quality captions for 45.8\% of test images, despite being trained on only half the data volume of FT w/ mixed data. This significant improvement validates the effectiveness of HMGIE-based data cleansing in enhancing model performance.
Qualitative analysis through representative examples in Figure \ref{fig:table15} reveals distinct characteristics of each model variant.
The original model tends to generate concise and generally accurate captions but often misses important details. The model fine-tuned with mixed data introduces more details but also risks generating hallucinated or inaccurate descriptions. In contrast, the model fine-tuned with cleaned data  achieves an optimal balance, generating comprehensive captions that maintain semantic accuracy while capturing nuanced visual details. These results empirically demonstrate the effectiveness of HMGIE as a data cleansing framework for vision-language tasks.

\begin{algorithm*}[!t]
\caption{Hierarchical and Multi-Grained Inconsistency Evaluation}
\label{alg:hmgie}
\SetKwInOut{Input}{Input}
\SetKwInOut{Output}{Output}
\DontPrintSemicolon

\Input{Image $I$, caption $T$, maximum level $L$}
\Output{H-Scores ($\mathcal{H}_{acc}$, $\mathcal{H}_{comp}$), consistency decision $d$, explanation $\xi$}

Generate semantic graph $\mathcal{G}_s$ from caption $T$\;

Initialize coverage mask $\mathcal{M}^0$ to all ones\;

Initialize empty HIEG $\mathcal{G}_{hie}^0$\;

\For{$l \leftarrow 1$ \KwTo $L$}{
    Generate $(\mathcal{Q}^l, \mathcal{A}_{ref}^l, \mathcal{R}^l)$ using semantic graph $\mathcal{G}_s$, mask $\mathcal{M}^{l-1}$, and previous HIEG $\mathcal{G}_{hie}^{l-1}$ \tcp*{Eq. (2)}
    
    \ForEach{question $q_i^l \in \mathcal{Q}^l$}{
        Generate VQA answer and confidence $(a_{vqa,i}^l, c_i^l)$ using $f_{vqa}$\;
        
        Evaluate answer correctness $y_i^l \in \{0,1\}$ using $f_{eval}$\;
    }
    
    Construct new node set $\mathcal{N}^l$ from evaluation quartets $(q_i^l, a_{ref,i}^l, a_{vqa,i}^l, y_i^l)$\;
    
    Expend HIEG with new nodes: $\mathcal{G}_{hie}^l \leftarrow \mathcal{G}_{hie}^{l-1} \cup \mathcal{N}^l \cup \mathcal{R}^l$ \tcp*{Eq. (3)}
    
    Update $\mathcal{M}^l$ by checking the semantic coverage between $\mathcal{G}_{hie}^l$ and $\mathcal{G}_s$   \tcp*{Eq. (4)}
    \If{$\mathcal{M}^l = \mathbf{0}$ }{  
        break; \tcp*{All elements in $\mathcal{G}_s$ are covered}
    }
}

Calculate accuracy score $\mathcal{H}_{acc}$ and completeness score $\mathcal{H}_{comp}$ \tcp*{Eq. (5) \& (6)}

Calculate consistency decision $d$ from all $y_i^l$ values\;

Generate natural language explanation $\xi$ from final HIEG $\mathcal{G}_{hie}^L$\;

\Return{$\mathcal{H}_{acc}$, $\mathcal{H}_{comp}$, $d$, $\xi$}\;
\end{algorithm*}

\begin{table*}[htbp]\centering
\caption{The prompt used to directly prompt MLLMs (DP) for visual-textual inconsistency detection.}
\label{tab:prompt_dp}
\begin{minipage}{\linewidth}
\vspace{0mm}   
\centering
\begin{tcolorbox} 
\small
\hspace{-6mm}

Does the image match the given caption? Answer with ``Yes'' or ``No'' with explanation. 

\ 

Caption: 

[\myred{\textbf{\{caption\}}}]

\ 

The output should be in JSON format:

\{

\qquad``Answer'': ``Yes" or ``No",

\qquad``Explanation'': ``Explanation of the answer"

\}
\end{tcolorbox}
\end{minipage}
\end{table*}

\begin{table*}[htbp]\centering
\caption{The prompt used to prompt MLLMs with chain-of-thought (CoT) for visual-textual inconsistency detection.}
\label{tab:prompt_cot}
\begin{minipage}{\linewidth}
\vspace{0mm}   
\centering
\begin{tcolorbox} 
\small
\hspace{-6mm}

You are a meticulous assistant specialized in detecting image-text consistency. Your task is to analyze whether the image matches the given caption.  Please do not overlook any slight discrepancy and evaluate the image-text consistency \underline{step by step}. Answer with ``Yes'' or ``No'' with explanation.

\ 

Caption: 

[\myred{\textbf{\{caption\}}}]

\

The output should be in JSON format:

\{

\qquad``Answer'': ``Yes" or ``No",

\qquad``Explanation'': ``Explanation of the answer"

\}
\end{tcolorbox}
\end{minipage}
\end{table*}

\begin{table*}[htbp]\centering
\caption{The prompt used to semantic graph generation.}
\label{tab:prompt_sg}
\begin{minipage}{\linewidth}
\vspace{0mm}   
\centering
\begin{tcolorbox} 
\small
\hspace{-6mm}

Your task is to parse the given image description (Caption) into a Semantic Graph. This semantic graph should capture the key entities, relationships, and attributes, and any other information in the description. Please follow these guidelines:

\ 

1. Node Types:

\qquad- Entity: People, animals, objects, etc.

\qquad- Location: Scenes or places

\qquad- Concept: Abstract Concept

\qquad- Event: Actions or events that occur

\qquad- Attribute: Characteristics describing entities, locations, or events

\qquad- Others

\ 

2. Edge Types:

  \qquad- Action: Actions performed by a subject on an object
  
  \qquad- Spatial: Representing the spatial relationship between entities and locations.
  
  \qquad- Has Attribute: Features of entities, locations, or events
  
  \qquad- Part Of: Representing compositional relationships
  
  \qquad- Quantity: Representing amounts or counts
  
  \qquad- Others

\ 

3. Output JSON Format:

\{

  \qquad``nodes": [
  
  \qquad\qquad\{``id": ``N1", ``type": ``Entity", ``label": ``entity name"\},
    
  \qquad\qquad\{......\}
  
  \qquad],
  
  \qquad``edges": [
  
   \qquad\qquad\{
    
      \qquad\qquad\qquad``from": (``N1", ``entity-name-1"),
      
      \qquad\qquad\qquad``to": (``N3", ``entity-name-3"),
      
      \qquad\qquad\qquad``type": ``Action",
      
      \qquad\qquad\qquad``label": ``action description",
      
      \qquad\qquad\qquad``description": ``A sentence describing this triple relationship"
      
   \qquad\qquad\},
    
    \qquad\qquad\{......\}
    
  \qquad]
  
\}

\ 

Please ensure:

\qquad- Each node has a unique ID (N1, N2, N3...)

\qquad- The ``from" and ``to" fields in edges use node IDs and labels

\qquad- Capture all important information in the Caption, but avoid over-inferring details that are not present

\qquad- Each edge includes a ``description" field with a sentence describing the triple relationship

\ 

Here is an example:
[\myred{\textbf{\{example\}}}]

\ 

Now, please parse the following Caption into a semantic graph:

[\myred{\textbf{\{caption\}}}]

\end{tcolorbox}
\end{minipage}
\end{table*}

\begin{table*}[htbp]\centering
\caption{The prompt used to question and reference generation.}
\label{tab:prompt_qg}
\begin{minipage}{\linewidth}
\vspace{0mm}   
\centering
\begin{tcolorbox} 
\small
\hspace{-6mm}

You are a specialized question generator for fine-grained semantic inconsistency detection.
Your task is to generate questions based on the semantic graph and previous hierarchical inconsistency evaluation graph (HIEG) for the current level to verify the semantic consistency between an image and its caption.
Focus on generating questions only for the current level, progressively increasing the difficulty and granularity to capture easily overlooked details as the level grows.

\ 

The input includes:

1. Semantic Graph (in JSON format):

   \qquad- Nodes: entities and concepts mentioned in the caption
   
   \qquad- Edges: relationships between nodes
   
2. Previous HIEG, where each node contains:

   \qquad - Question-ID: unique identifier
   
   \qquad - Question: the actual question text
   
   \qquad - Verify-Fact: the fact that this question is trying to verify
   
   \qquad - Expected-Answer: answer derived from semantic graph
   
   \qquad - Actual-Answer: answer provided by VQA module
   
   \qquad - Eval-Correct: Whether the Actual-Answer is correct. True is correct, False is 
   incorrect.
   
   \qquad - Parent-IDS: IDs of questions this question depends on
   
3. Current Level: the depth for new questions to be generated

4. Suggestion:  guide the direction or focus for this level of questions

\ 

You need to generate different new questions for the current level. Each question should:

1. Build upon the history HIEG without repetition.

2. Generate the questions based on the semantic graph. Use the semantic graph for reference.

3. Explore unverified nodes and edges in the semantic graph that are not covered by the history HIEG.

4. Increase in difficulty and granularity compared to the previous level, but not excessively.

5. Consider the direction or points raised in the suggestion, but not totally depend on it.

6. Avoid generating questions that involve vague or relative attributes (e.g., ``Is the pottery large?” or ``Is the object small?”). Questions should not require answers based on subjective sizes or undefined comparisons.

\ 

Please Ensure questions align with the depth appropriate for the current level.
The levels of questions are as follows: 

-Level 1: Essential Scene Elements;
Primary objects/subjects identification;
Basic scene setting;

-Level 2: Primary Details \& Relationships; Basic attributes of main objects (colors, sizes, basic states); Simple relationships between primary objects; Basic actions, events and interactions

-Level 3: Supporting Elements \& Complex Details; Secondary objects and their basic attributes; More detailed attributes of primary objects; More complex spatial relationships

-Level 4: Fine-grained Details \& Complex Relationships
-- Subtle attributes and characteristics
-- Complex multi-object relationships
-- Detailed spatial arrangements

-Level 5 and above: Comprehensive Scene Understanding
-- Highly specific object details
-- Subtle variations in attributes
-- Precise spatial configurations
-- Environmental nuances
-- Logical consistency across all elements

\ 

Output should be in JSON format as follows:

...

\ 

The input are:

Semantic graph:[\myred{\textbf{\{semantic-graph\}}}]

History HIEG:[\myred{\textbf{\{previous-HIEG\}}}]

Suggestion: [{\myred{\textbf{\{suggestion\}}}}]

Current Level: [\myred{\textbf{\{current-level\}}}]

\end{tcolorbox}
\end{minipage}
\end{table*}

\begin{table*}[htbp]\centering
\caption{The prompt used to visual question answering.}
\label{tab:promt_vqa}
\begin{minipage}{\linewidth}
\vspace{0mm}   
\centering
\begin{tcolorbox} 
\small
\hspace{-6mm}

You are an advanced AI assistant specialized in visual question answering.

The question is: [\myred{\textbf{\{question\}}}]

\ 

Examine the question and the image carefully, analyze the image in detail and provide an accurate and answer step by step. Then provide a brief explanation if necessary. Provide a confidence score for your answer on a scale of 0 to 1, where 0 indicates low confidence and 1 indicates absolute certainty.

\ 
Output in JSON Format:

\{

  \qquad``Answer": ``Your answer here",
  
  \qquad``Confidence": X.X

\}

\end{tcolorbox}
\end{minipage}
\end{table*}

\begin{table*}[htbp]\centering
\caption{The prompt used to question evaluation.}
\label{tab:promt_qe}
\begin{minipage}{\linewidth}
\vspace{0mm}   
\centering
\begin{tcolorbox} 
\small
\hspace{-6mm}

You are a question answer evaluator. 
Your task is to evaluate if actual answers is correct.
You are given a visual question about an image, an expected answer, and an actual answer.
Since the question is about an image, the actual answer may be more specific than the expected answer.
The answer is correct as long as the actual answer is semantically similar to the expected answer, or contains more specific details that include the expected answer, or contains most key elements but missing minor details.
The answer is incorrect if there is explicit contradiction with expected answer, or the answer is completely unrelated.

\ 

Output should be in JSON format:

\{

    \qquad``Correct": boolean
    
\}

\ 

The Input:

Question: [\myred{\textbf{\{question\}}}]

Expected Answer: [\myred{\textbf{\{expected-answer\}}}]

Actual Answer: [\myred{\textbf{\{actual-answer\}}}]

\end{tcolorbox}
\end{minipage}
\end{table*}

\begin{table*}[htbp]\centering
\caption{The prompt used to semantic coverage checking.}
\label{tab:promt_verify}
\begin{minipage}{\linewidth}
\vspace{0mm}   
\centering
\begin{tcolorbox} 
\small
\hspace{-6mm}

You are an AI assistant responsible for detecting image-text inconsistency based on QA results. 
The questions are verifying the facts in the given Semantic Graph for an image's text description.
Existing question-answer pairs form a hierarchical inconsistency evaluation graph (HIEG). As the level increases, the questions become more specific.
Your task is checking whether existing HIEG have verified all the semantically important elements in the semantic graph, providing suggestions for next question generation if needed. 

\ 

Input:

1. Semantic graph parsed from the text description of an image.

2. Current HIEG, each node containing:

  \qquad- Question-ID: A unique identifier.
  
  \qquad- Question: The actual question text.
  
  \qquad- Verify-Fact: The fact that this question is trying to verify.
  
  \qquad- Expected-Answer:  The answer expected if the image and text were match.
  
  \qquad- Actual-Answer: Real received answers based on the actual image.
  
  \qquad- Evaluation-Correct: Whether the Actual-Answer is correct. True is correct, False is incorrect.
  
  \qquad- Parent-IDs: IDs of questions that this question depends on or follows from.

\ 

Your task is to check verification completeness: Examine if all important elements from the Semantic Graph have been covered by the questions in the HIEG, regardless of the correctness of the answers.  Consider verification complete if all important elements have been verified, and set Next-Level-Suggestion to None. If there are still unverified important elements, provide specific suggestions for the next level of questions to verify remaining key elements.

\ 

Please note that when suggesting next level suggestions,
  a. Only suggest questions for uncovered elements from the semantic graph
  b. Do not repeat previously asked questions..
  c. The suggestions should be the direction of the next level of questions, not the specific questions.

\ 

The levels of questions in the HIEG are defined as follows:

-Level 1: Essential Scene Elements;
Primary objects/subjects identification;
Basic scene setting;

-Level 2: Primary Details \& Relationships; Basic attributes of main objects (colors, sizes, basic states); Simple relationships between primary objects; Basic actions, events and interactions

-Level 3: Supporting Elements \& Complex Details; Secondary objects and their basic attributes; More detailed attributes of primary objects; More complex spatial relationships

-Level 4: Fine-grained Details \& Complex Relationships
-- Subtle attributes and characteristics
-- Complex multi-object relationships
-- Detailed spatial arrangements

-Level 5 and above: Comprehensive Scene Understanding
-- Highly specific object details
-- Subtle variations in attributes
-- Precise spatial configurations
-- Environmental nuances
-- Logical consistency across all elements

\ 

Please timely stop suggesting questions if verification is complete.

Output should be JSON format:

...

\       

The input information are as follows:

Semantic Graph: [\myred{\textbf{\{semantic-graph\}}}]

HIEG: [\myred{\textbf{\{hieg\}}}]

\end{tcolorbox}
\end{minipage}
\end{table*}

\begin{table*}[htbp]\centering
\caption{The prompt used to explanation generation.}
\label{tab:promt_explain}
\begin{minipage}{\linewidth}
\vspace{0mm}   
\centering
\begin{tcolorbox} 
\small
\hspace{-6mm}

You are an AI assistant responsible for generating natural language explanations for visual-textual consistency evaluation results. Based on the Hierarchical Inconsistency Evaluation Graph (HIEG) and the final consistency decision, you will provide clear, structured explanations that trace the evaluation process and highlight key findings.

\ 

Input:

1. HIEG structure where each node contains:

  \qquad- Question-ID: A unique identifier.
  
  \qquad- Question: The actual question text.
  
  \qquad- Verify-Fact: The fact that this question is trying to verify.
  
  \qquad- Expected-Answer:  The answer expected if the image and text were match.
  
  \qquad- Actual-Answer: Real received answers based on the actual image.
  
  \qquad- Evaluation-Correct: Whether the Actual-Answer is correct. True is correct, False is incorrect.
  
  \qquad- Parent-IDs: IDs of questions that this question depends on or follows from.

2. Original Caption: The text description being evaluated

3. Final Consistency Decision: ``Consistent" or ``Inconsistent"

\ 

Your task is to generate a comprehensive explanation that:

1. For Inconsistent Cases:

   \qquad- Start with a clear statement of inconsistency
   
   \qquad- Present inconsistencies in a hierarchical order (from basic to detailed)
   
   \qquad- For each inconsistency:Identify the specific semantic element involved; Explain the discrepancy between the caption and image; Reference the relevant evaluation level and question ID
     
   \qquad- Highlight relationships between related inconsistencies
   
   \qquad- Provide a concise summary of the impact on overall semantic meaning

2. For Consistent Cases:

   \qquad - Begin with a confirmation of consistency
   
   \qquad - Summarize the key semantic elements verified
   
   \qquad - Highlight important relationships and attributes confirmed
   
   \qquad - Organize verification results by evaluation levels
   
   \qquad - Emphasize any notable detailed verifications
   
   \qquad - Conclude with overall semantic alignment confirmation
   
\ 

Guidelines for Explanation Generation:

\qquad - Maintain a clear progression through evaluation levels

\qquad - Use precise language to describe semantic relationships

\qquad - Connect related findings across different levels

\qquad - Highlight the granularity of verified details

\qquad - Keep explanations concise but informative

\qquad - Use natural, flowing language while maintaining technical accuracy

\ 

Output Format:
...

Input:

HIEG: [\myred{\textbf{\{hieg\}}}]

Original Caption: [\myred{\textbf{\{caption\}}}]

Final Consistency Decision:[\myred{\textbf{\{consistency-decision\}}}]

\end{tcolorbox}
\end{minipage}
\end{table*}

\begin{table*}[!h]
    \centering
    \caption{Examples labeled as consistent but detected as inconsistent by HMGIE in NewsCLIPpings~\cite{luo2021newsclippings}}
    \label{tab:fpr_examples_newsclip}
    \renewcommand{\arraystretch}{0.4}
    \setlength{\tabcolsep}{6pt}
    \small
    \begin{tabular}{>{\centering\arraybackslash}m{5cm} >{\raggedright\arraybackslash}m{5cm} >{\raggedright\arraybackslash}m{5cm} >{\raggedright\arraybackslash}p{5cm}}
        \toprule
        \textbf{Image} & \textbf{Caption} & \textbf{Explanation} \\
        \midrule
        \includegraphics[width=4cm]{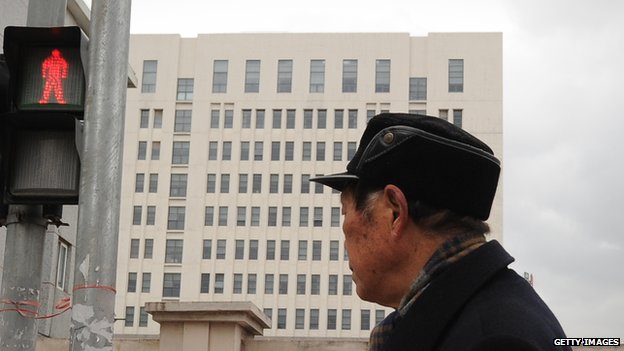} & The US says some Chinese hackers work in this Shanghai building. & No direct visual evidence or context linking the person, the building, or the scene to hacking activities or Shanghai hackers. \\
        \midrule
        \includegraphics[width=4cm]{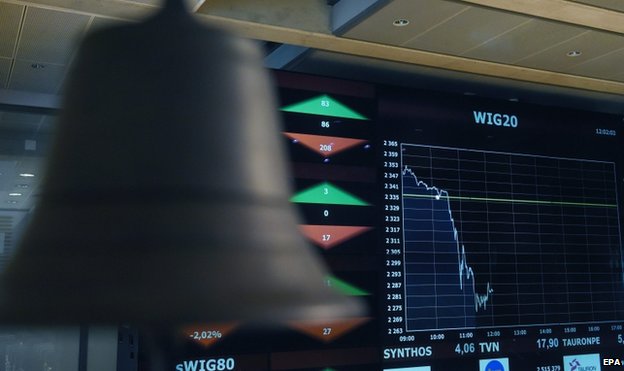} & Shares plummeted in Warsaw and the Swiss franc soared against the Zloty after the Swiss cap ended. & No direct semantic correlation between the image and the text. \\
        \midrule
        \includegraphics[width=4cm]{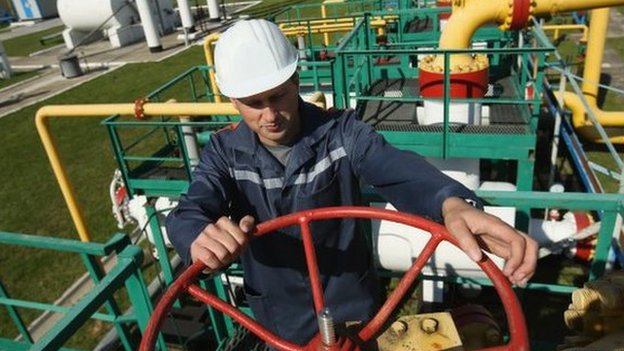} & Ukraine relies on gas imports from western Europe but storage is low ahead of the winter season. & No direct semantic correlation between the image and the text. \\
        \bottomrule
    \end{tabular}
\end{table*}

\begin{table*}[h]
    \centering
    \caption{Examples labeled as consistent but detected as inconsistent by HMGIE in TIIL~\cite{huang2024dtiil})}
    \label{tab:fpr_examples_tiil}
    \renewcommand{\arraystretch}{0.4}
    \setlength{\tabcolsep}{6pt}
    \small
    \begin{tabular}{>{\centering\arraybackslash}m{5cm} >{\raggedright\arraybackslash}m{5cm} >{\raggedright\arraybackslash}m{5cm} >{\raggedright\arraybackslash}p{5cm}}
        \toprule
        \textbf{Image} & \textbf{Caption} & \textbf{Explanation} \\
        \midrule
        \includegraphics[width=4cm]{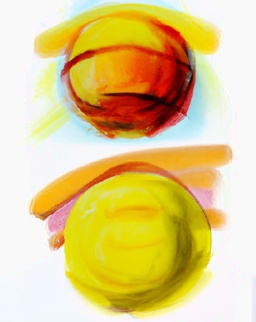} & The artist who created the sketches said they weigh 80 pounds each and required four months to create. & No direct semantic correlation between the image and the text. \\
        \midrule
        \includegraphics[width=4cm]{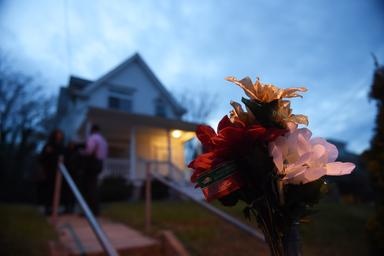} & Flowers were placed outside Barry's home. & Requires out-of-context knowledge to verify. \\
        \midrule
        \includegraphics[width=4cm]{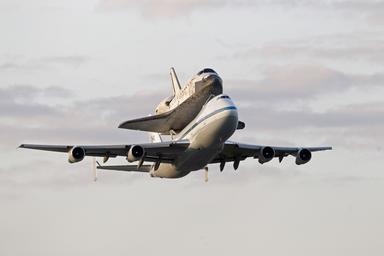} & View Photo Gallery: Space shuttle Discovery landed at Dulles airport on April 17 on the back of a 747 after a flyaround of much of the region. It is now bound for the Smithsonian's Steven F Udvar-Hazy Center. & The text provides much more information than the image, causing a misalignment. \\
        \bottomrule
    \end{tabular}
\end{table*}

\begin{table*}[h]
    \centering
    \caption{Examples labeled as consistent but detected as inconsistent by HMGIE in SeeTRUE~\cite{yarom2024you}}
    \label{tab:fpr_examples_seetrue}
    \renewcommand{\arraystretch}{0.4} 
    \setlength{\tabcolsep}{6pt}
    \begin{tabular}{>{\centering\arraybackslash}m{5cm} >{\raggedright\arraybackslash}p{5cm} >{\raggedright\arraybackslash}m{5cm} >{\raggedright\arraybackslash}p{5cm}}
        \toprule
        \textbf{Image} & \textbf{Caption} & \textbf{Explanation} \\
        \midrule
        \includegraphics[width=4cm]{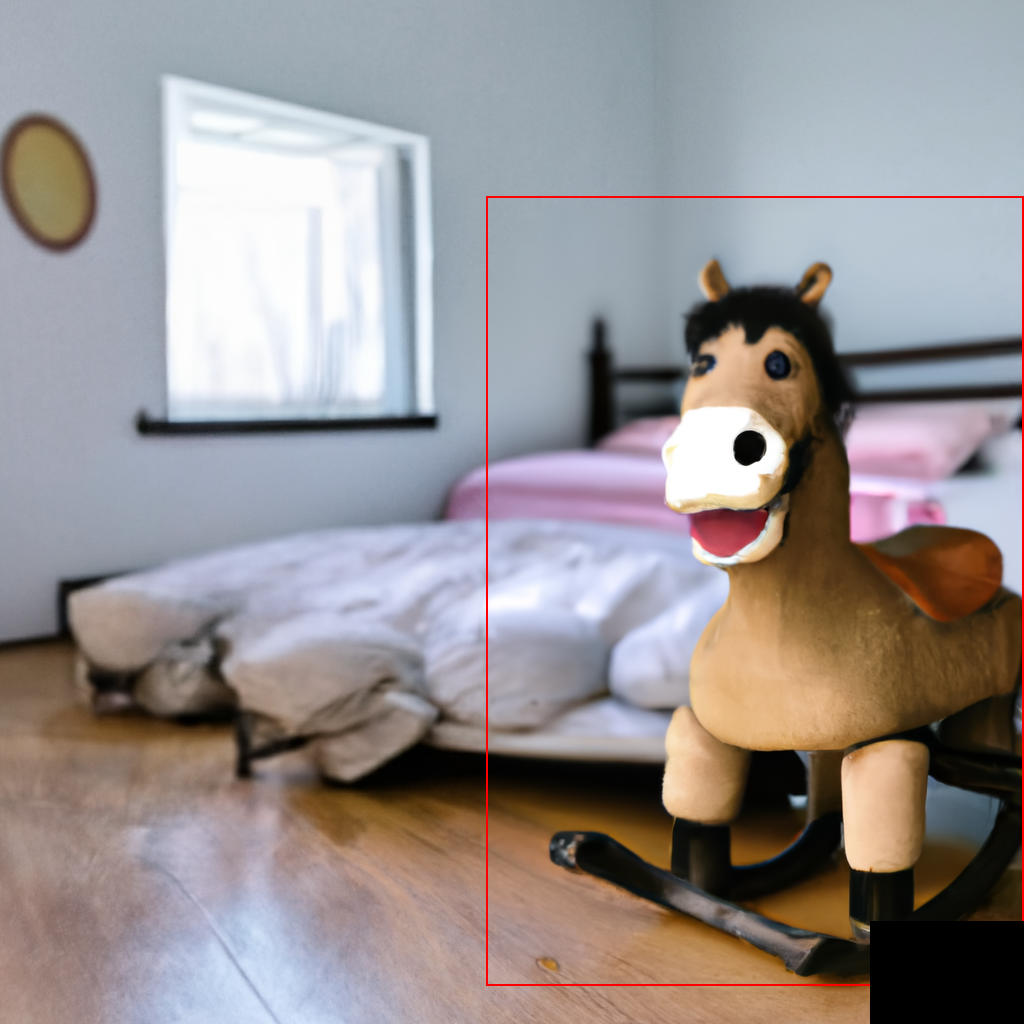} & A cloth horse sitting in a bedroom with many big beds. & Only one bed in the bedroom, not many big beds. \\
        \midrule
        \includegraphics[width=4cm]{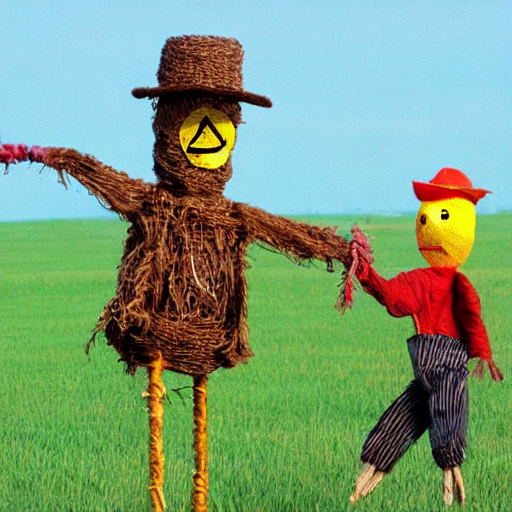} & A bird scaring a scarecrow. & There is no bird present. \\
        \midrule
        \includegraphics[width=4cm]{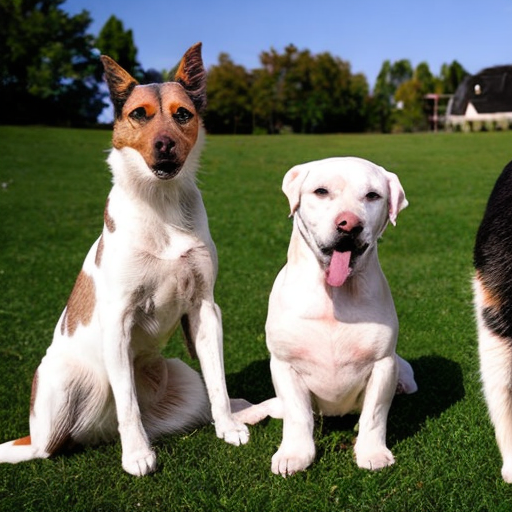} & One cat and two dogs sitting on the grass. & There is no visible cat present. \\
        \bottomrule
    \end{tabular}
\end{table*}

\end{document}